\documentclass{article}
\usepackage{times}
\usepackage{epsfig}
\usepackage{graphicx}
\usepackage{subfigure}
\usepackage{amsmath}
\usepackage{amsmath,amsthm}
\usepackage{algorithmic}
\usepackage{algorithm}
\usepackage{authblk}
\PassOptionsToPackage{numbers, compress}{natbib}
\usepackage[preprint]{neurips_2019}
\usepackage{xcolor}
\usepackage{caption}
\usepackage{wrapfig}

\newtheorem{theorem}{Theorem}
\newtheorem{lemma}{Lemma}

\newtheorem{remark}{Remark}

\newcommand*{\ShowNotes}{}

\ifdefined\ShowNotes
  \newcommand{\colornote}[3]{{\color{#1}\bf{#2: #3}\normalfont}}
\else
  \newcommand{\colornote}[3]{}
\fi

\newfont{\msym}{msbm10}

\newcommand{\paren}[1]{\left({#1}\right)}

\newcommand{\braces}[1]{\left\{{#1}\right\}}

\newcommand{\Exp}[1]{{\rm E}\left[{#1}\right]}

\newcommand{\beq}[1]{\begin{equation}\label{#1}}
\newcommand{\eeq}{\end{equation}}
\newcommand{\beqa}{\begin{eqnarray}}
\newcommand{\eeqa}{\end{eqnarray}}

\newcommand{\mb}[1]{{\boldsymbol{#1}}}

\newcommand{\vv}{\mb{v}}

\newcommand{\vomega}{\mb{\omega}}

\newcommand{\vw}{\vomega}

\newcommand{\newstufffroma}[1]{}

\newcommand{\newstufffrom}[1]{}

\newcommand{\oldnote}[2]{}

\newcommand{\comment}[1]{}
\newcommand{\commentout}[1]{}

\newcounter {mySubCounter}
\newcommand {\twocoleqn}[4]{
  \setcounter {mySubCounter}{0} %
  \let\OldTheEquation \theequation %
  \renewcommand {\theequation }{\OldTheEquation \alph {mySubCounter}}%
  \noindent \hfill%
  \begin{minipage}{.40\textwidth}
\vspace{-0.6cm}
    \begin{equation}\refstepcounter{mySubCounter}
      #1
    \end {equation}
  \end {minipage}
~~~~~~
  \addtocounter {equation}{ -1}%
  \begin{minipage}{.40\textwidth}
\vspace{-0.6cm}
    \begin{equation}\refstepcounter{mySubCounter}
      #3
    \end{equation}
  \end{minipage}%
  \let\theequation\OldTheEquation
}

\usepackage[utf8]{inputenc} 
\usepackage[T1]{fontenc}    
\usepackage{hyperref}       
\usepackage{url}            
\usepackage{booktabs}       
\usepackage{amsfonts}       
\usepackage{nicefrac}       
\usepackage{microtype}      
\title{XNAS: Neural Architecture Search \\with Expert Advice}

\author{%
  \textbf{Niv Nayman$^*$, Asaf Noy$^*$, Tal Ridnik\thanks{These authors contributed equally.} , Itamar Friedman, Rong Jin, Lihi Zelnik-Manor} \\
  Machine Intelligence Technology, Alibaba Group \\
  \texttt{\{niv.nayman,asaf.noy,tal.ridnik,itamar.friedman,jinrong.jr,lihi.zelnik\}}
  \\\texttt{@alibaba-inc.com} \\
}

\begin{document}
\maketitle

\begin{abstract}
This paper introduces a novel optimization method for differential neural architecture search, based on the theory of prediction with expert advice. Its optimization criterion is well fitted for an architecture-selection, i.e., it minimizes the regret incurred by a sub-optimal selection of operations. 
Unlike previous search relaxations, that require hard pruning of architectures, our method is designed to dynamically wipe out inferior architectures and enhance superior ones.
It achieves an optimal worst-case regret bound and suggests the use of multiple learning-rates, based on the amount of information carried by the backward gradients. 
Experiments show that our algorithm achieves a strong performance over several image classification datasets.
Specifically, it obtains an error rate of $1.6\%$ for CIFAR-10, $24\%$ for ImageNet under mobile settings, and achieves state-of-the-art results on three additional datasets. 

\end{abstract}

\section{Introduction}
In recent years tremendous efforts have been put into a manual design of high performance neural networks~\cite{larsson2016fractalnet,hu2018squeeze,szegedy2016rethinking_label_smooth,szegedy2015going}.
An emerging alternative approach is replacing the manual design with automated Neural Architecture Search (\textbf{NAS}).
NAS excels in finding architectures which yield state-of-the-art results. 
Earlier NAS works were based on reinforcement learning~\cite{zoph2016neural,NASNET}, sequential optimization~\cite{PNAS}, and evolutionary algorithms~\cite{Real18Regularized}, and required immense computational resources, sometimes demanding years of GPU compute time in order to output an architecture.
More recent NAS methods reduce the search time significantly, e.g. via weight-sharing~\cite{ENAS} or by a continuous relaxation of the space~\cite{liu2018darts}, making the search affordable and applicable to real problems.

While current NAS methods provide encouraging results, they still suffer from several shortcomings.
For example, a large number of hyper-parameters that are not easy to tune, hard pruning decisions that are performed sub-optimally at once at the end of the search, and a weak theoretical understanding. 
This cultivates skepticism and criticism of the utility of NAS in general.
Some recent works even suggest that current search methods are only slightly better than random search and further imply that some selection methods are not well principled and are basically random~\cite{li2019random,sciuto2019evaluating}.

To provide more principled methods, we view NAS as an online selection task, and rely on  \emph{Prediction with Experts Advice} (\textbf{PEA}) theory \cite{cesa2006prediction} for the selection.
Our key contribution is the introduction of XNAS (eXperts Neural Architecture Search), an optimization method (\autoref{sec:arch_search}) that is well suited for optimizing inner architecture weights over a differentiable architecture search space (\autoref{sec:nueral_architecture_space}).
We propose a setup in which the experts represent inner neural operations and connections, whose dominance is specified by architecture weights. 

Our proposed method addresses the mentioned shortcomings of current NAS methods. For the mitigation of the hard pruning, we leverage the Exponentiated-Gradient (EG) algorithm \cite{kivinen1997exponentiated}, which favors sparse weight vectors to begin with, enhanced by a wipeout mechanism for dynamically pruning inferior experts during the search process. 
Additionally, the algorithm requires less hyper-parameters to be tuned (\autoref{sec:no_weight_decay}), and the theory behind it further provides guidance for the choice of learning rates. Specifically, the algorithm avoids the decay of architecture weights \cite{goodfellow2016deep}, which is shown to promote selection of arbitrary architectures.

Additionally, XNAS features several desirable properties, such as achieving an optimal worst-case regret bound (\autoref{sec:theoretical_analysis}) and suggesting to assign different learning rates for different groups of experts. Considering an appropriate reward term, the algorithm is more robust to the initialization of the architecture weights and inherently enables the recovery of 'late bloomers', i.e., experts which may become effective only after a warm-up period (\autoref{sec:recovery_of_late_bloomers}). The wipeout mechanism allows the recovery of experts with a chance of being selected at the end of the process.

We compare XNAS to previous methods and demonstrate its properties and effectiveness over statistical and deterministic setups, as well as over $7$ public datasets (\autoref{sec:Experiments and Results}). It achieves state-of-the-art performance over $3$ datasets, and top-NAS over rest, with significant improvements. For example, XNAS reaches $1.60\%$ error over CIFAR-10, more than $20\%$ improvement over existing NAS methods.

\section{Proposed Approach}


To lay out our approach we first reformulate the differentiable architecture search space of DARTS~\cite{liu2018darts} in a way that enables direct optimization over the architecture weights.
We then propose a novel optimizer that views NAS as an online selection task, and relies on PEA theory for the selection.

\subsection{Neural Architecture Space}
\label{sec:nueral_architecture_space}


We start with a brief review of the PEA settings and then describe our view of the search space as separable PEA sub-spaces. This enables us to leverage PEA theory for NAS. 

\textbf{PEA Settings.} PEA~\cite{cesa2006prediction} refers to a sequential decision making framework, dealing with a decision maker, i.e. a forecaster, whose goal is to predict an unknown outcome sequence $\{y_t\}_{t=1}^T \in \mathcal{Y}$ while having access to a set of $N$ experts' advises, i.e. predictions. Denote the experts' predictions at time $t$ by $f_{t,1},\dots, f_{t,N}\in \mathcal{D}$, where $\mathcal{D}$ is the \emph{decision space}, which we assume to be a convex subset of a vector space. Denote the forecaster's prediction $\{p_t\}_{t=1}^T \in \mathcal{D}$, and a non-negative loss function $\ell:\mathcal{D}\times\mathcal{Y}\xrightarrow{}\mathbb{R}$. 
At each time step $t=1,\dots,T$, the forecaster observes $f_{t,1},\dots, f_{t,N}$ and predicts $p_t$. The forecaster and the experts suffer losses of $\ell_t(p_t):=\ell(p_t , y_t)$ and $\ell_t(f_{t,i}):=\ell(f_{t,i}, y_t)$ respectively.

\textbf{The Search Space Viewed as Separable PEA Sub-spaces.}
We view the search space suggested by DARTS~\cite{liu2018darts} as multiple separable sub-spaces of experts, as illustrated in Figure \ref{fig:forecaster}, described next.
An architecture is built from replications of \emph{normal} and \emph{reduction cells} represented as a directed acyclic graph. 
Every node $x^{(j)}$ in the graph represents a feature map and each directed edge $(j, k)$ is associated with a forecaster, that predicts a feature map $p^{(j,k)}:= p^{(j,k)}(x^{(j)})$ given the input $x^{(j)}$.
Intermediate nodes are computed based on all of their predecessors: $x^{(k)} =\Sigma_{j<k}p^{(j,k)}$. The output of the cell is obtained by applying a reduction operation (e.g. concatenation) to the intermediate nodes.
During the search stage, every forecaster combines $N$ experts' feature map predictions $\{f_i^{(j,k)}\}_{i=1}^N:=\{f_i^{(j,k)}(x^{(j)})\}_{i=1}^N$ forming its own prediction,
\begin{align}\label{eq:forecaster}
p^{(j,k)}=\sum_{i=1}^N u_i^{(j,k)}f_i^{(j,k)} \quad ; \quad u^{(j,k)}_i=\frac{v^{(j,k)}_i}{\sum_{l=1}^N v^{(j,k)}_l} \quad ; \quad v^{(j,k)}_i \geq 0
\end{align}
From now on, we will ignore the superscript indices $(j,k)$ for brevity. Each expert represents a neural operation, e.g. convolution or max pooling, associated with network weights $w_{i}$, that receives an input $x_t$ at time $t$, and outputs a feature map prediction. Thus a time index is attached, as each prediction $f_{t,i}:=f_i(x_t)$ is associated with updated weights $w_{t,i}$.

Our architecture search approach is composed of two stages.
In the \emph{search stage}, the weights $w_{t,i}$ and $v_{t,i}$ are alternately optimized as described in~\autoref{sec:arch_search}; then, in the \emph{discretization stage}, a discrete child architecture is obtained as explained next.

\textbf{The Discretization Stage.} \label{sec:pruning_stage}
Once the architecture weights are optimized, the final discrete neural architecture is obtained by performing the following discretization stage, adopted from \cite{liu2018darts}: Firstly, the strongest two predecessor edges are retained for each intermediate node. The strength of an edge is defined as $\max_{i}{u_i^{(k,j)}}$.
Lastly, every forecaster is replaced by the corresponding strongest expert.

\subsection{XNAS: eXperts Neural Architecture Search}
\label{sec:arch_search}

The differential space, described in section \ref{sec:nueral_architecture_space}, enables direct optimization over the architecture weights via gradient-descent based techniques. Previous methods adopted generic optimizers commonly used for training the network weights. For example~\cite{liu2018darts,xie2018snas,chen2019progressive,casale2019probabilistic} used adam \cite{kingma2014adam}, and~\cite{noy2019asap} used SGD with momentum. 
While those optimizers excel in joint minimization of neural network losses when applied to network weights, NAS is a essentially a \emph{selection} task, aiming to select a subset of experts out of a superset. The experts weights form a convex combination, as they compete over a forecaster's attention.

We argue that a generic alternate optimization of network weights and architecture weights, as suggested in previous works, e.g. \cite{hundt2019sharpdarts,liu2018darts}, is not suitable for the unique structure of the architecture space. Hence, we design a tailor-made optimizer for this task, inspired by PEA theory. In order to evaluate experts' performance, a loss is to be associated with each expert. However, an explicit loss is not assigned to each expert, as opposed to a back-propagated loss gradient.
Therefore, we base our algorithm on a version of the Exponentiated-Gradient (EG) algorithm adapted for the NAS space. EG algorithms favor sparse weight vectors~\cite{kivinen1997exponentiated}, thus fit well to online selection problems. 

We introduce the XNAS (eXperts Neural Architecture Search) algorithm, outlined in Algorithm~\ref{alg:XNAS} for a single forecaster.
XNAS alternates between optimizing the network weights $\mb{w}$ and the architecture weights $\mb{v}$ in a designated manner. After updating $\mb{w}$ (line~\ref{alg_w_update}) by descending the train loss $\ell_{train}$, the forecaster makes a prediction based on the mixture of experts (line~\ref{alg_forward}), then $\mb{v}$ are updated with respect to the validation loss $\ell_{val}$ according to the EG rule (lines \ref{alg_rewards}-\ref{update_reward2}). 
Next, the optimizer wipes-out weak experts (lines~\ref{wipe_rule}-\ref{eq:wipeout}), effectively assigning their weights to the remaining ones. 
The exponential update terms, i.e. rewards, are determined by the projection of the loss gradient on the experts' predictions: $R_{t,i}=-\nabla_{p_t}\ell_{\mathrm{val}} (p_t)\cdot f_{t,i}$. 
In section \ref{sec:theoretical_analysis} we relate to the equivalence of this reward to the one associated with \emph{policy gradient} search \cite{williams1992simple} applied to NAS.

\noindent\begin{minipage}[b]{.5\textwidth}
\begin{algorithm}[H] 
    \caption{XNAS for a single forecaster} 
    \label{alg:XNAS}
    \begin{algorithmic}[1]
        \STATE \textbf{Input}: The learning rate $\eta$,  \\    
        Loss-gradient bound $\mathcal{L}$ \label{grad_bound}, \\ 
        Experts predictions $\left\{f_{t,i}\right\}_{i=1}^N\, \forall t=1,\dots,T$
        \STATE \textbf{Init}: $I_0 = \{1, \dots,N\},~ v_{0,i} \leftarrow 1, ~\forall i\in I_0$
        \FOR{rounds $t=1,\dots,T$}
        \STATE Update $\vw$ by descending $\nabla_{\vw}\ell_{\mathrm{train}}(\vw,\mb{v})$ \label{alg_w_update}
        \STATE $p_t \leftarrow \frac{\sum_{i\in I_{t-1}} v_{t-1,i} \cdot f_{t-1,i}}{\sum_{i\in I_{t-1}} v_{t-1,i}}$~~~~~~~~~~\text{\#Predict} \label{alg_forward}
        \STATE $\braces{\text{loss gradient revealed: }\nabla_{p_t}\ell_{\mathrm{val}} (p_t)}$
        \FOR{$i\in I_{t-1}$}
        \STATE $R_{t,i}=-\nabla_{p_t}\ell_{\mathrm{val}} (p_t)\cdot f_{t,i}$ ~~~~\text{\#Rewards}\label{alg_rewards} \label{update_reward}
        \STATE $v_{t,i} \leftarrow v_{t-1,i}\cdot \exp{\{\eta R_{t,i} \}}$ ~~~\text{\#EG step} \label{update_reward2}
        \ENDFOR
        \STATE $\theta_t \leftarrow \max\limits_{i\in I_{t-1}}\{v_{t,i}\} \cdot \exp{\{-2\eta\mathcal{L}(T-t)\}}$ \label{wipe_rule}
        \STATE $I_t \leftarrow I_{t-1} \setminus \left\{i \mid  v_{t,i} < \theta_t\right\}$  ~~~~~~~\text{\#Wipeout} \label{eq:wipeout}
        \ENDFOR
    \end{algorithmic}
\end{algorithm}

\end{minipage}
\begin{minipage}[b]{.5\textwidth}
    \centering                
    \includegraphics[width=\linewidth, height=180pt]{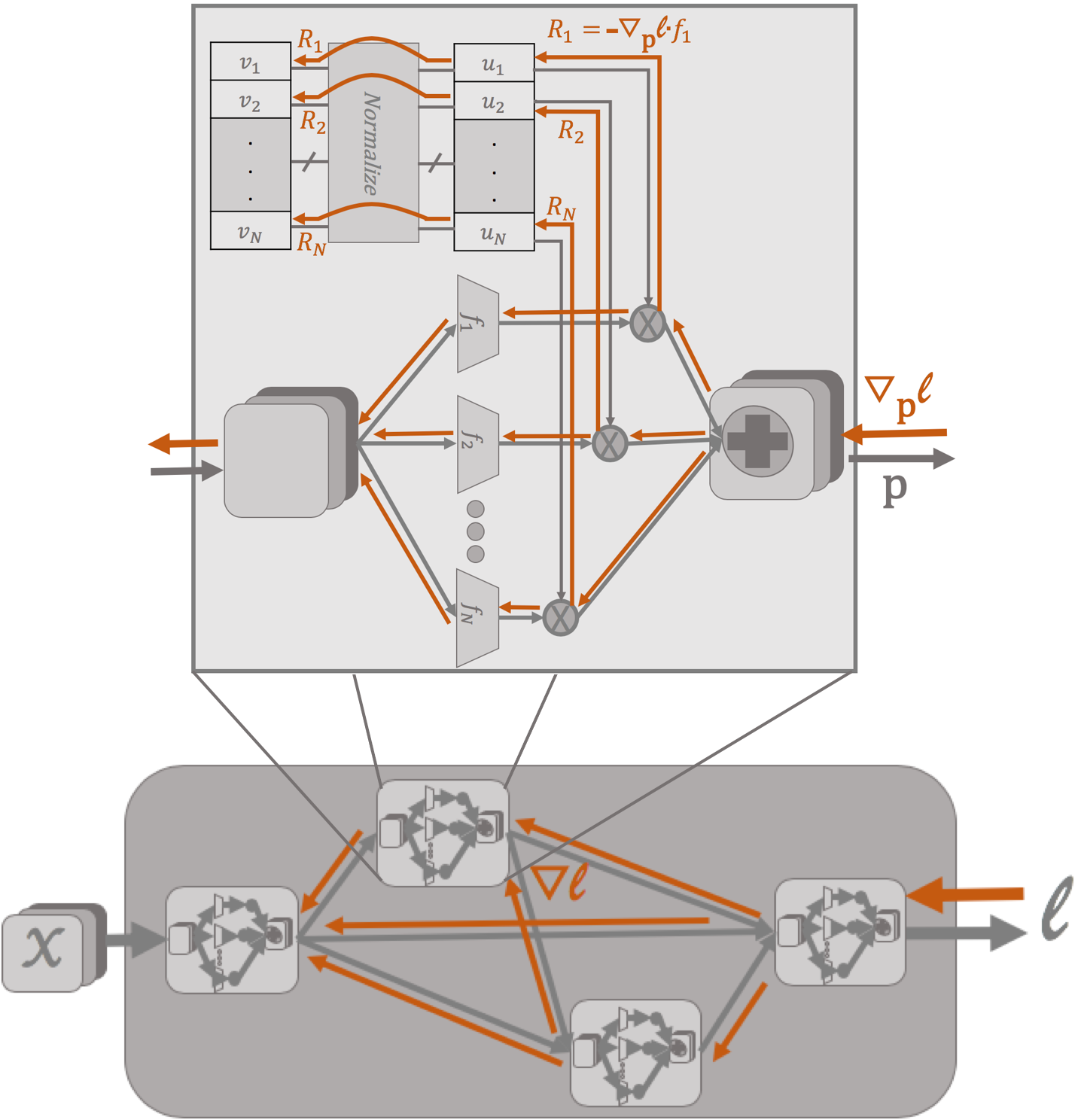}
    \captionof{figure}{A visual illustration  of the search space as separable experts sub-problems, equipped with XNAS forward and backward propagation. forward and backward passes are presented as grey and orange arrows respectively.} 
    \label{fig:forecaster}
\end{minipage}

The purpose of the wipeout step is threefold. First, the removal of weak experts consistently smooths the process towards selecting a final architecture at the descretization stage described in section \ref{sec:nueral_architecture_space}. Thus it mitigates the harsh final pruning of previous methods, which results in a \emph{relaxation bias} addressed in ~\cite{snas, noy2019asap}. Second, it dynamically reduces the number of network weights, thus simplifying the optimization problem, avoiding over-fitting and allowing it to converge to a better solution. Last, it speeds-up the architecture search, as the number of graph computations decreases with the removal of experts.   

\section{Analysis and Discussion}
\subsection{Theoretical Analysis}
\label{sec:theoretical_analysis}
In this section we analyse the performance of the proposed algorithm. For this purpose, we introduce the \emph{regret} as a performance measure for NAS algorithms. Showing that the wipeout mechanism cannot eliminate the best expert (lemma \ref{no_wipe}), we provide theoretical guarantees for XNAS with respect to that measure (theorem \ref{EGW_Optimality}). Proofs appear in Section \ref{subsec:proofs} of the supplementary material for brevity. 
Relaying on the theoretical analysis, we extract practical instructions with regard to the choice of multiple learning rates. Finally we briefly discuss the equivalence of our reward to the one considered by the policy gradient approach applied to NAS.

\textbf{The Regret as a Performance Measure.}
Denote the regret, the cumulative losses of the forecaster and of the $i$th expert at time $t$ by,
\begin{align}
\label{losses_and_regret}
\textrm{Regret}_t = L_t-\min_{i=1,\dots N}L_{t,i}   \quad ; \quad 
 L_t=\sum_{s=1}^{t}\ell_s(p_s) \quad ; \quad  L_{t,i} = \sum_{s=1}^t \ell_s(f_{s,i})
\end{align} 
respectively. The regret measures how much the forecaster regrets not following the advice of the best expert in hindsight. This criterion suits our setup as we optimize a mixture of experts and select the best one by the end of the process.

In classical learning theory, statistical properties of the underlying process may be estimated on the basis of stationarity assumptions over the sequence of past observations. Thus effective prediction rules can be derived from these estimates~\cite{cesa2006prediction}. However, NAS methods that alternately learn the architecture weights and train the network weights are highly non-stationary.
In PEA theory, no statistical assumptions are made, as ``simplicity is a merit''~\cite{hazan2016introduction}, and worst-case bounds are derived for the forecaster's performance. We obtain such bounds for the wipeout mechanism and the regret. 

\textbf{A Safe Wipeout.} In XNAS (line \ref{wipe_rule}), by the choice of the wipeout thresholds, experts with no chance of taking the lead by the end of the search are wiped out along the process. In a worse-case setup, a single incorrect wipeout might result in a large regret, i.e., linear in the number of steps $T$, due to a loss gap at each consecutive step. The following lemma assures that this cannot happen,
\begin{lemma}
\label{no_wipe}
In XNAS, the optimal expert in hindsight cannot be wiped out. 
\end{lemma}
The wipeout effectively transfers the attention to leading experts. Define the \emph{wipeout factor} and the \emph{aggregated wipeout factor} as $\Gamma_t := 1 + \frac{\sum_{i \in I_{t-1}\setminus I_t} v_{t,i}}{\sum_{i\in I_t}  v_{t,i}}$ and $\gamma_T :=\prod_{t=1}^T\Gamma_t$, respectively.
\begin{lemma}
\label{lemma:gamma_bound}
The aggregated wipeout factor satisfies $1 \leq \gamma_T < N$.
\end{lemma}
Equipped with both lemmas, we show that the wipeout may improve the EG regret bound for certain reward sequences. 

\textbf{Regret Bounds.} Our main theorem guarantees an upper bound for the regret, 
\begin{theorem}[XNAS Regret Bound]
\label{EGW_Optimality}
The regret of the XNAS algorithm \ref{alg:XNAS}, with $N$ experts and learning rate $\eta$, incurring a sequence of $T$ non-negative convex losses of $\mathcal{L}$-bounded rewards, satisfies,
\begin{align}\label{eq:regret_bound}
\mathrm{Regret}_T \leq \frac{\eta T \mathcal{L}^2}{2} +\frac{1}{\eta}\ln N -\frac{1}{\eta}\ln{\gamma_T}
\end{align}
\end{theorem}
As an input parameter of XNAS, the learning rate $\eta$ cannot be determined based on the value of $\gamma_T$, since the later depends on the data sequence. Choosing the minimizer $\eta^*$ of the first two terms of \eqref{eq:regret_bound} fully known in advance, yields the following tight upper bound,
\begin{align}
\label{eq:eta_and_regret}
\eta^*=\sqrt{\frac{2\ln N}{T\mathcal{L}^2}} \quad ; \quad \mathrm{Regret}_T  \leq \mathcal{L}\sqrt{2T\ln N}\paren{1 - \frac{1}{2} \frac{\ln{\gamma_T}}{\ln{N}}} 
\end{align}
The regret upper bound of XNAS is tight, as the lower bound can be shown to be of $\Omega(\mathcal{L}\sqrt{T\ln{N}})$~\cite{haussler1995tight}. In addition, the wipeout related term reduces the regret in an amount which depends on the data sequences through $\gamma_T$, as it effectively contributes the attention of weak experts to the leading ones. For comparison, under the same assumptions, the worst-case regret bound of gradient-descent is of $O(\mathcal{L}\sqrt{T N})$~\cite{hazan2016introduction}, while the one of Adam is linear in $T$~\cite{reddi2019convergence}. 
An illustration of the relationship between the regret and the rate of correct expert selection appears in section \ref{sec:stochastic_toy} of the supplementary material, where XNAS is shown to achieve a better regret compared to a generic optimizer.\\

\textbf{Multiple Learning Rates.} Equation \ref{eq:eta_and_regret} connects the optimal theoretical learning rate $\eta^*$ with the number of steps $T$, which is also the number of gradient feedbacks received by the experts. Since forecasters weights are being replicated among different cells, the number of feedbacks is different for normal and reduction cells (\autoref{sec:nueral_architecture_space}). Explicitly, $T_c = d \times \mathcal{E} \times r_c$, where $T_c,d,\mathcal{E},r_c$ are the effective horizon $T$, the validation set size, the number of epochs and the number of replications for cell type $c$ respectively. We adopt the usage of multiple learning rates $\eta^*_c$ in our experiments as upper bounds on the learning rates for minimizing the upper bound of the regret.
\\

\textbf{The Connection to Policy Gradient.} We conclude this section by pointing out an interesting connection between \emph{policy gradient} in NAS \cite{zoph2016neural} and PEA theory. We refer to the PEA based reward term in line~\ref{update_reward} of algorithm \ref{alg:XNAS}. This reward has been shown by \cite{snas} to be the same effective reward of a policy gradient method applied to the common NAS optimization criterion~\cite{zoph2016neural, NASNET,ENAS}. More precisely, consider the case where instead of mixing the experts' predictions, the forecaster is to sample a single expert at each step with probability $u_{t,i}$, specified in \eqref{eq:forecaster}. Then the effective reward associated with the policy gradient will be exactly the derived PEA reward, $R_{t,i}=-\nabla_{p_t}\ell_t (p_t)\cdot f_{t,i}$. XNAS optimizes with respect to the same reward, while avoiding the sampling inefficiency associated with policy gradient methods.

\subsection{Key Properties and Discussion}
In this section we discuss some of the key properties of XNAS. For each of this properties we provide supporting derivations, illustrations and demonstrations appearing in section \ref{sec:key_properties_supports} of the supplementary material for brevity.
\vspace{-0.1cm}
\subsubsection{The Recovery of Late Bloomers and Robustness to Initialization}
\label{sec:recovery_of_late_bloomers}
In this section we point out a key difference between our proposed update rule and the one used in previous works. 
We refer to Gradient Descent (GD) with softmax as updating the parameters $\alpha_{t,i} = \ln{v_{t,i}}$, by descending $\nabla_{\mb{\alpha_{t,i}}}\ell_t(p_t)$ respectively.
Variants of GD with softmax, as used in \cite{liu2018darts} to optimize the architecture weights, suppress operations that are weak at the initial iterations, making it more difficult for them to ``bloom'' and increase their weights. This could be problematic, e.g. in the two following cases.
First, consider the best expert starting with a poor weight which gradually rises. This could be the case when an expert representing a parameterized operation (e.g. a convolutional layer) competes with an unparameterized one (e.g. a pooling layer), as the first requires some period for training, as stated by \cite{noy2019asap,hundt2019sharpdarts}.
 Second, consider a noisy setup, where the best expert in hindsight could receive some hard penalties before other inferior experts do. In NAS we deal with stochastic settings associated with the training data.

We inspect the update term of GD with softmax,
\begin{align}
   v_{t,i} = \exp\{\alpha_{t,i}\} = \exp\{\alpha_{t-1,i} - \eta\partial_{\alpha_{t-1,i}}\ell_t(p_t)\} = v_{t-1,i}\cdot\exp\{-\eta \partial_{\alpha_{t-1,i}}\ell_t(p_t)\}
\end{align}
Hence, the effective reward in this case is,
\begin{align} \label{GD_SM_update}
\tilde{R}_{t,i}:= -\partial_{\alpha_{t-1,i}}\ell_t(p_t) = -\nabla_{p_t}\ell_t(p_t)\cdot u_{t-1,i}\left(f_{t,i}-p_t\right)
\end{align}
See derivations in section \ref{sec:gradients_derivations}. The linear dependence on the expert's weight $u_{t-1,i}$ in \eqref{GD_SM_update} implies that GD with softmax makes it harder for an expert whose weight is weak at some point to recover and become dominant later on, as the associated rewards are attenuated by the weak expert's weight.

XNAS mitigates this undesirable behavior.
Since for XNAS the update term \eqref{update_reward} depends on the architecture weights only indirectly, i.e. through the prediction, the recovery of \emph{late bloomers} is not discouraged, as demonstrated in section \ref{sec:deterministic_toy} of the supplementary material. From the very same reasons, XNAS is more robust to the initialization scheme compared to GD with softmax and its variants, as demonstrated in section \ref{sec:2d_toy} of the supplementary material. These advantages make XNAS more suitable for the NAS setup. 

Note that while the XNAS enables the recovery of experts with badly initialized weights or with delayed rewards, the wipeout mechanism prevents inferior operations that start blooming too late from interfering, by eliminating experts with no chance of leading at the end.

\textbf{Wipeout Factor.} As mentioned in section~\ref{sec:arch_search}, the wipeout mechanism contributes to both optimization process and search duration. A further reduction in duration can be achieved when the wipe-out threshold in line \ref{wipe_rule} of Algorithm 1 is relaxed with a parameter $0< \zeta \leq 1$, being replaced by $\theta_t \leftarrow \max_{i\in I_{t-1}}\{v_{t,i}\} \cdot \exp{\{{-2\eta\mathcal{L}(T-t)\cdot\zeta\}}}$. This will lead to a faster convergence to a single architecture, with the price of a violation of the theoretical regret. As worst-case bounds tend to be over pessimistic, optimizing over $\zeta$ could lead to improved results. We leave that for future work.   

\vspace{-0.2cm}
\subsubsection{Fewer Hyper Parameters}
\vspace{-0.2cm}
\label{sec:no_weight_decay}
The view of the differentiable NAS problem as an optimization problem solved by variants of GD, e.g. Adam, introduces some common techniques for such schemes along with their corresponding hyper-parameters. Tuning these complicates the search process - the fewer hyper-parameters the better.
We next discuss how XNAS simplifies and reduces the number of hyper-parameters.

\textbf{Theoretically Derived Learning Rates.}
The determination of the learning rate has a significant impact on the convergence of optimization algorithms. Various scheduling schemes come up, e.g. \cite{loshchilov2016sgdr,smith2017cyclical}, as the later additionally suggests a way for obtaining an empirical upper bound on the learning rate. In section \ref{sec:theoretical_analysis}, multiple learning rates $\eta^*_c$ are suggested for minimizing the regret bound \eqref{eq:eta_and_regret}, as $c\in\{N, R\}$ represents normal and reduction cells respectively.
 For example, for CIFAR10 with 50\%:50\% train-validation split, 50 search epochs, gradient clipping of $1$, $6$ normal cells and $2$ reduction cells both of $8$ experts for each forecaster, \eqref{eq:eta_and_regret} yields $\eta^*_N=$7.5e-4 and $\eta^*_R=$1.3e-3.
\begin{remark} Note that the proposed learning rates minimize an upper bound of the regret \eqref{eq:eta_and_regret} in the case of no wipeout, i.e. the worst case, as the extent of the wipeout cannot be known in advance. Hence the proposed learning rate provides an upper bound on the optimal learning rates and can be further fine-tuned.
\end{remark}
\begin{wrapfigure}{r}{0.5\textwidth}
    \vspace{-0.4cm}
    \captionof{figure}{Mean normalized entropy vs weight decay. The red dot refers to DARTS' settings.}
    \centering
    \includegraphics[width=0.5\textwidth]{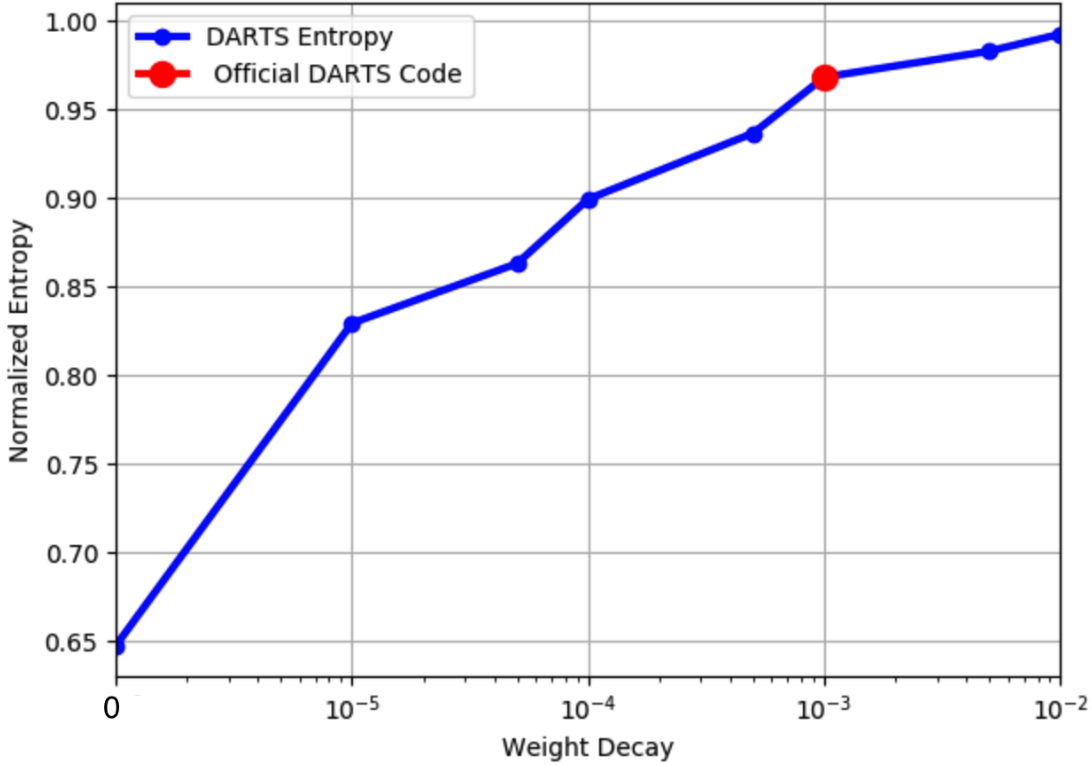}
    \label{fig:entropy_vs_wd}
    \vspace{-0.7cm}
\end{wrapfigure}

\vspace{-0.1cm}
\textbf{No Weight Decay.}
Another common  technique involving hyper-parameters is weight decay, which has no place in the theory behind XNAS.
We claim that the obviation of weight decay by XNAS makes sense. Regularization techniques, such as weight decay, reduce over-fitting of over-parametrized models when applied to these parameters \cite{goodfellow2016deep}. No such effect is incurred when applying weight decay on the architecture parameters
as they do not play the same role as the trained network parameters $\mb{w}$. Instead, weight decay encourages uniform dense solutions, as demonstrated in Figure \ref{fig:entropy_vs_wd}, where the mean normalized entropy increases with the weight decay coefficient. The calculation of the mean normalized entropy is detailed in section \ref{sec:mean_normalized_entropy} of the supplementary material.
This observation could be associated with the suggestion of recent works \cite{li2019random,sciuto2019evaluating} that current search methods are only slightly better than random search.
The density of results in a harder degradation in performance once \emph{discretization stage} occurs (section \ref{sec:nueral_architecture_space}), hence sparse solutions are much preferred over dense ones. 
   
\textbf{No Momentum.} 
The theory behind XNAS obviates momentum \cite{qian1999momentum} and ADAM's exponentially decay rates \cite{kingma2014adam}. Since momentum requires more state variables and more computations, the resulting XNAS optimizer turns out to be simpler, faster and with a smaller memory footprint, compared to commonly used optimizers for NAS, e.g. ADAM \cite{liu2018darts,xie2018snas,chen2019progressive,casale2019probabilistic} and SGD with momentum \cite{noy2019asap}.





\section{Experiments and Results}
\label{sec:Experiments and Results}
In this section we will test XNAS on common image classification benchmarks, and show its effectiveness compared to the other state-of-the-art models.

We used the CIFAR-10 dataset for the main search and evaluation phase. In addition, using the cell found on CIFAR-10 we did transferability experiments on the well-known benchmarks ImageNet, CIFAR-100, SVHN, Fashion-MNIST, Freiburg and CINIC10.

\vspace{-0.2cm}
\subsection{Architecture Search on CIFAR-10} \label{arch_search_cifar}
\vspace{-0.2cm}
Using XNAS, we searched on CIFAR-10 in a small parent network for convolutional cells. 
Then we built a larger network by stacking the learned cells, trained it on CIFAR-10 and compared the results against other NAS methods.

We created the parent network by stacking $8$ cells with $4$ ordered nodes, each of which connected via forecasters to all previous nodes in the cell and also to the two previous cells outputs.
Each forecaster contains seven operations: 3x3 and 5x5 separable and dilated separable convolutions, 3x3 max-pooling, 3x3 average-pooling and an identity.
A cells output is a concatenation of the outputs of the four cells nodes.

The search phase lasts $50$ epochs. We use the first-order approximation \cite{liu2018darts}, relating to \( \vv \) and \( \vw \) as independent parameters which can be optimized separately. The train set is divided into two parts of equal sizes: one is used for training the operations weights $\vw$ and the other for training the architecture weights $\vv$. With a batch size of $96$, one epoch takes $8.5$ minutes in average on a single GPU\footnote{Experiments were performed using a NVIDIA GTX 1080Ti GPU.}, summing up to $7$ hours in total for a single search. Figure~\ref{fig:cells} shows our learned normal and reduction cells, respectively.



\vspace{-0.2cm}
\subsection{CIFAR-10 Evaluation Results}
\label{CIFAR-10-Evaluation-Results}
\vspace{-0.3cm}
\begin{table}[H]
\begin{minipage}[t]{.5\linewidth}
    We built the evaluation network by stacking $20$ cells: $18$ normal cells and $2$ reduction cells. The reduction cells are placed after $1/3$ and $2/3$ of the network. After each reduction cell we double the amount of channels in the network. We trained the network for $1500$ epochs using a batch size of $96$ and SGD optimizer with nesterov-momentum.
    Our learning rate regime was composed of $5$ cycles of power cosine annealing learning rate~\cite{hundt2019sharpdarts}, with amplitude decay factor of $0.5$ per cycle. For regularization we used cutout~\cite{devries2017improved}, scheduled drop-path~\cite{larsson2016fractalnet}, auxiliary towers~\cite{szegedy2015going}, label smoothing~\cite{szegedy2016rethinking_label_smooth} AutoAugment~\cite{cubuk2018autoaugment} and weight decay. 
    To understand the effect of the network size on the final accuracy, we chose to test $3$ architecture configurations, XNAS-Small, XNAS-Medium and XNAS-Large, with $36$, $44$ and $50$ initial network channels respectively. Table~\ref{cifar-10-results} shows the performance of XNAS architectures compared to other state-of-the-art NAS methods.
    We can see from Table~\ref{cifar-10-results} that XNAS smallest network variant, XNAS-Small, outperforms on CIFAR-10 previous NAS methods by a large margin.
    Our largest network variant, XNAS-Large, is the second highest reported score on CIFAR-10 (without additional pre-train data), while having $4$ times less parameters than top one~\cite{cubuk2018autoaugment}. In addition, XNAS is among the fastest NAS methods.
\end{minipage}
\quad
\begin{minipage}[t]{.45\linewidth}
    \vspace{-0.2cm}
    \centering
    \begin{tabular}[t]{|l|c|c|c|}
    \hline
    \vtop{\hbox{\strut \textbf{CIFAR-10}}\hbox{\strut Architecture}}
    & \multicolumn{1}{|p{0.75cm}}{\centering Test \\error}
    & \multicolumn{1}{|p{0.9cm}}{\centering Params\\(M) }
    & \multicolumn{1}{|p{0.8cm}|}{\centering Search\\cost}\\
    \hline
    AutoAugment \cite{cubuk2018autoaugment}  & $1.48$  & $26$  & $20$ \\    
    \hline
    NAONet-WS \cite{NAO} & $3.53$ & $2.5$ & $0.3$ \\   			
    PNAS \cite{PNAS} & $3.41$  & $3.2$  & $150$ \\  			
    Amoeba-A \cite{Real18Regularized}     & $3.34$  & $3.2$  & $3150$ \\
    DSO-NAS \cite{DSO} & $2.95$ & $3$ & $1$ \\
    DARTS(1nd) \cite{liu2018darts}      & $2.94$       & $2.9$  & $0.4$ \\
    ENAS \cite{ENAS} & $2.89$  & $4.6$  & $0.5$ \\
    SNAS \cite{snas} & $2.85$ & $2.8$ & $1.5$ \\
    DARTS(2nd) \cite{liu2018darts} & $2.67$  & $3.4$  & $1$ \\
    NASNet-A \cite{NASNET}  & $2.65$      & $3.3$  & $1800$ \\			
    Amoeba-B \cite{Real18Regularized} & $2.55$ & $2.8$  & $3150$ \\
    PDARTS \cite{chen2019progressive} & $2.50$ & $3.4$ & $0.3$ \\      
    NAONet \cite{NAO} & $2.11$ & $128$ & $200$ \\  
    ProxylessNAS \cite{cai2018proxylessnas} & $2.08$ & $5.7$ & 10 \\  
    ASAP + ET \cite{noy2019asap} & $1.94$ & $3.7$ & $0.2$ \\
    SharpDarts \cite{hundt2019sharpdarts} & $1.93$ & $3.6$ & $0.8$ \\       
    \hline
    XNAS-Small & $\bf{1.81}$ & $3.7$ & $0.3$ \\
    XNAS-Medium & $\bf{1.73}$ & $5.6$ & $0.3$ \\
    XNAS-Large & $\bf{1.60}$ & $7.2$ & $0.3$ \\
    \hline
    \end{tabular}
\captionof{table}{Classification errors of XNAS compared to state-of-the-art methods on CIFAR-10. The Search cost is measured in GPU days. Test error refers to top-1 test error (\%). XNAS Small, Medium and Large refers to network architectures with XNAS cell and $36$, $44$ and $50$ initial channels respectively.}
\label{cifar-10-results}
\end{minipage}
\end{table}

\subsection{Transferability Evaluation}\label{Transfer learning results}
\begin{table}[H]
\begin{minipage}[H]{.5\linewidth}
    \vspace{-0.5cm}
    Using the cell found by XNAS  search on CIFAR-10, we preformed transferability tests on $6$ popular classification benchmarks: ImageNet, CIFAR-100, Fashion-MNIST, SVHN, Freiburg and CINIC10. \\
    
    \vspace{-0.2cm}
    {\bf {ImageNet Results.}}
    Our ImageNet network was composed of $14$ stacked XNAS cells, with two initial stem cells for downscaling. We used $46$ initial channels so the total number of network FLOPs is below 600[M], similar to other ImageNet architectures with small computation regime~\cite{xie2019exploring}. We trained the network for $250$ epochs with one cycle of power cosine learning rate and a nesterov-momentum optimizer. 
    The results are presented in Table~\ref{imagenet-results}. We can see from Table~\ref{imagenet-results} that XNAS transferability results on ImageNet are highly competitive, outperforming all previous NAS cells.
\end{minipage}
\quad
\begin{minipage}[H]{.45\linewidth}
    \vspace{-0.2cm}
    \centering
    \begin{tabular}{|l|c|c|c|}
    \hline
    \vtop{\hbox{\strut \textbf{ImageNet}}\hbox{\strut Architecture}}
    & \multicolumn{1}{|p{0.8cm}}{\centering Test\\error }
    & \multicolumn{1}{|p{0.9cm}}{\centering Params\\(M) }            
    & \multicolumn{1}{|p{0.8cm}|}{\centering Search\\cost}\\
    \hline
    SNAS \cite{snas}  & $27.3$ & $4.3$ & $1.5$\\ 
    ASAP \cite{noy2019asap}  & $26.7$ & $5.1$ & $0.2$\\
    DARTS \cite{liu2018darts} 
    & $26.7$ & $4.9$
    & $1$\\
    NASNet-A \cite{NASNET} & $26.0$ & $5.3$ & $1800$\\
    PNAS \cite{PNAS}  & $25.8$ & $5.1$ & $150$\\   
    Amoeba-A \cite{Real18Regularized}  &  $25.5$ & $5.1$ & $3150$\\
    RandWire \cite{xie2019exploring} & $25.3$ & $5.6$ & $0$ \\
    SharpDarts \cite{hundt2019sharpdarts} & $25.1$ & $4.9$ & $0.8$ \\    
    Amoeba-C \cite{Real18Regularized}  & $24.3$ & $6.4$ & $3150$\\
    \hline
    XNAS & $\bf{24.0}$ & $5.2$ & 0.3\\
    \hline
\end{tabular}
\captionof{table}{Transferability classification error of XNAS, compared to top NAS cells, on ImageNet. Test error refers to top-1 test error (\%). Search cost is measured in GPU days.}
\label{imagenet-results}
\end{minipage}
\end{table}

\vspace{-0.5cm}
{\bf {Additional Results.}}
We further tested XNAS transferability abilities on $5$ smaller datasets: CIFAR-100 \cite{cifar100}, Fashion-MNIST \cite{fashionMnist}, SVHN \cite{SVHN}, Freiburg \cite{Freiburg} and CINIC10 \cite{darlow2018cinic}. We chose to use the XNAS-Small architecture, with similar training scheme to the one described in section \ref{CIFAR-10-Evaluation-Results}. Table~\ref{cifar-100-results} shows the performance of our model compared to NAS methods. We can see that XNAS cell excels on the datasets tested. On CIFAR-100 it surpasses the next top cell by $1$\%, achieving the second highest reported score on CIFAR-100 (without additional pre-train data), second only to \cite{cubuk2018autoaugment}. On Fashion-MNIST, Freiburg and CINIC10, to the best of our knowledge XNAS achieves a new state-of-the-art accuracy.
\vspace{-0.1cm}
\begin{table}[H]
    \begin{center}
        \begin{tabular}{|l|c|c|c|c|c|c|c|}
        \hline
        \vtop{\hbox{\strut \textbf{Datasets}}\hbox{\strut Architecture}}
            & \multicolumn{1}{|p{1.4cm}}{\centering CIFAR100 Error}
            & \multicolumn{1}{|p{1.2cm}}{\centering FMNIST Error}
            &
            \multicolumn{1}{|p{1.1cm}}{\centering SVHN Error}
            &\multicolumn{1}{|p{1.2cm}}{\centering Freiburg Error}
            &\multicolumn{1}{|p{1.2cm}}{\centering CINIC10 Error}
            & \multicolumn{1}{|p{1.0cm}}{\centering Params\\(M) }
            & \multicolumn{1}{|p{0.8cm}|}{\centering Search\\cost}\\
            \hline
            Known SotA 
            & $\bf{10.7}$ \cite{cubuk2018autoaugment} 
            & $3.65$ \cite{zhong2017random} 
            & $\bf{1.02}$ \cite{cubuk2018autoaugment} 
            & $10.7$ \cite{noy2019asap} 
            & $6.83$ \cite{noy2019asap}
            &26 \cite{cubuk2018autoaugment}
            & - \\
            \hline
             PDARTS \cite{chen2019progressive} & $15.9$ & - & - & - & - & $3.6$ & $0.3$ \\     
            NAONet-1 \cite{NAO} & $15.7$ & - & - & - & - & $10.8$ & $200$ \\       
            NAONet-2 \cite{NAO} & $14.7$ & - & - & - & - & $128$ & $200$ \\               
             PDARTS-L \cite{chen2019progressive} & $14.6$ & - & - & - & - & $11$ & $0.3$ \\   
            \hline
            SNAS$^\dagger$ \cite{snas} & $16.5$ & $3.72$ & $1.98$ & $14.7$ & $7.13$ & $2.8$ & $1.5$ \\ 
            PNAS$^\dagger$ \cite{PNAS} & $15.9$ & $3.72$ & $1.83$& $12.3$ & $7.03$ &$3.4$ & $150$ \\
            Amoeba-A$^\dagger$ \cite{Real18Regularized} & $15.9$ & $3.8$ & $1.93$ & $11.8$ & $7.18$ & $3.2$ & $3150$ \\ NASNet$^\dagger$ \cite{NASNET} & $15.8$ & $3.71$ & $1.96$ & $13.4$  & $6.93$ & $3.3$ & $1800$ \\   
            DARTS$^\dagger$ \cite{liu2018darts} & $15.7$ & $3.68$ & $1.95$ & $10.8$ & $6.88$ & $3.4$ &  $1$ \\
            ASAP$^\dagger$ \cite{noy2019asap} & $15.6$ & $3.71$ & $1.81$ & $10.7$  & $6.83$ & $2.5$ & $0.2$ \\ 
            \hline
            XNAS-Small & $13.6$ & $\bf{3.64}$ & $1.72$ & $\bf{6.3}$ & $\bf{6.0}$& $3.7$ &  0.3 \\
           \hline
        \end{tabular}
    \end{center}
    \caption{Classification errors of XNAS, compared to state-of-the-art NAS methods, on several datasets. Error refers to top-1 test error (\%). The Search cost is measured in GPU days. Results marked with $^\dagger$ are taken from \cite{noy2019asap}, which tested and compared different cells on various datasets.}
    \label{cifar-100-results}
\end{table}

\section{Related Work}
Mentions of Experts in deep learning~\cite{rasmussen2002infinite,yao2009hierarchical, garmash2016ensemble, aljundi2017expert} literature go decades back~\cite{Jacobs:1991, chen1999improved}, typically combining models as separable experts sub-models. 
A different concept of using multiple mixtures of experts as inner parts of a deep model, where each mixture has its own gating network, is presented in \cite{eigen2013learning}. 
Following works build upon this idea and include a gating mechanism per mixture~\cite{CondConv2019, teja2018hydranets}, and some further suggest sparsity regularization over experts via the gating mechanism~\cite{shazeer2017outrageously, wang2018deep}.
These gating mechanisms can be seen as a dynamic routing, which activates a single or a group of experts in the network on a per-example basis. Inspired by these works, our methods leverage PEA principled methods for automatically designing neural network inner components.

Furthermore, optimizers based on PEA theory may be useful for the neural architecture search phase.
Common stochastic gradient-descent (SGD) and a set of PEA approaches, such as follow-the-regularized-leader (FTRL), were shown by~\cite{shalev2012online, hazan2016introduction, van2014follow} to be equivalent.
Current NAS methods~\cite{NASNET, zoph2016neural, ENAS, liu2018darts, cai2018proxylessnas, wu2018fbnet, li2019random, hundt2019sharpdarts, noy2019asap} use Adam, SGD with Momentum or other common optimizers.
One notion that is common in PEA principled methods is the regret~\cite{cesa2006prediction}. PEA strategies aim to guarantee a small regret under various conditions. We use the regret as a NAS objective, in order to establish a better principled optimizer than existing methods~\cite{li2019random,sciuto2019evaluating}. Several gradient-descent based optimizers, such as Adam, present a regret bound analysis, however, the worst-case scenario for Adam has non-zero average regret~\cite{reddi2019convergence}, i.e., it is not robust. Our optimizer is designated for selecting architecture weights while achieving an optimal regret bound.

\section{Conclusion}
\label{sec:conclusion}
In this paper we presented XNAS, a PEA principled optimization method for differential neural architecture search. 
Inner network architecture weights that govern operations and connections, i.e. experts, are learned via exponentiated-gradient back-propagation update rule.
XNAS optimization criterion is well suited for architecture-selection, since it minimizes the regret implied by sub-optimal selection of operations with tendency for sparsity, while enabling late bloomers experts to warm-up and take over during the search phase.
Regret analysis suggests the use of multiple learning rates based on the amount of information carried by the backward gradient. A dynamic mechanism for wiping out weak experts is used, reducing the size of computational graph along the search phase, hence reducing the search time and increasing the final accuracy. 
XNAS shows strong performance on several image classification datasets, while being among the fastest existing NAS methods.
\section*{Acknowledgements}
We would like to thank the members of the Alibaba Israel Machine Vision Lab (AIMVL), in particular to Avi Mitrani, Avi Ben-Cohen, Yonathan Aflalo and Matan Protter for their feedbacks and productive discussions. 

\newpage
\begin{center} \begin{huge}{Supplementary Material} \end{huge} \end{center}
\label{supplementary}
\section{Proofs}
\label{subsec:proofs}
\subsection{Proof of Lemma 1} 
\label{sec:wipe_lemma_proof}
\begin{proof}
By contradiction, assume that expert $j$ is being wiped-out at the iteration $t$ based on line~\ref{wipe_rule} in algorithm \ref{alg:XNAS}, and without loss of generality, $k\neq j$ is the leading expert at that time,
\begin{align} \label{contra}
v_{t,k} &= \arg\max_{i}\{v_{t,i}\} \notag \\
v_{t,j}&<v_{t,k} \cdot \exp{\{-2\eta\mathcal{L}(T-t)\}} 
\end{align}
Since expert $j$ is the optimal one in hindsight, specifically, 
\begin{align} 
\label{contra_hind}
v_{T,j} \geq v_{T,k}
\end{align}
However, since the loss is $\mathcal{L}$-bounded, the ratios between weights at time $t,T$ are bounded as well,
\begin{align} 
v_{T,k} &\geq v_{t,k} \cdot \exp{\{-\eta\mathcal{L}(T-t)\}}\\
v_{T,j} &\leq v_{t,j} \cdot \exp{\{\eta\mathcal{L}(T-t)\}}
\label{contra_bounded}
\end{align}
Recap inequalities \eqref{contra_hind}-\eqref{contra_bounded},
\begin{align*}
v_{t,j} &\geq  v_{T,j}\cdot\exp{\{-\eta\mathcal{L}(T-t)\}} 
\\ &\geq  
v_{T,k}\cdot\exp{\{-\eta\mathcal{L}(T-t)\}} 
\\ &\geq 
v_{t,k}\cdot\exp{\{-2\eta\mathcal{L}(T-t)\}}
\end{align*}
In contradiction to \eqref{contra}.
\qed
\end{proof}

\subsection{Proof of Lemma 2}
\begin{proof}
\textbf{Left Hand Side.} 
Due to the non-negativity of $\{v_{t,i}\}_{i=1}^N$ for all $t=1,\dots,T$, we have,
\begin{align}
    \Gamma_t = 1 + \frac{\sum_{i \in I_{t-1}\setminus I_t} v_{t-1,i}}{\sum_{i\in I_t}  v_{t,i}}
    \geq 1
\end{align}
Hence,
\begin{align}
    \gamma_T =\prod_{t=1}^T\Gamma_t \geq \prod_{t=1}^T 1 = 1
\end{align}

\textbf{Right Hand Side.}
Denote $|\mathcal{S}|$ as the cardinally of the set $\mathcal{S}$.
Then,
\begin{align}
    \Gamma_t 
    &= 
    1 + \frac{\sum_{i \in I_{t-1}\setminus I_t} v_{t,i}}{\sum_{i\in I_t}  v_{t,i}}
    \notag
    \\&<
    1 + \frac{|I_{t-1}\setminus I_t| \cdot \theta_t}{\sum_{i\in I_t}  v_{t,i}}
    \label{eq:wiped}
    \\&\leq
    1 + \frac{|I_{t-1}\setminus I_t| \cdot \theta_t}{|I_t| \cdot \theta_t}
    \label{eq:non_wiped}
    \\&=
    \frac{|I_t| + |I_{t-1}\setminus I_t|}{|I_t|}
    \notag
    \\&=
    \frac{|I_{t-1}|}{|I_t|}
    \notag
\end{align}
where \eqref{eq:wiped} is since $v_{t,i} < \theta_t \, \forall i\in I_{t-1}\setminus I_t$ and  \eqref{eq:non_wiped} is since $v_{t,i} \geq \theta_t \, \forall i\in I_t$, according to line \ref{eq:wipeout} of algorithm \ref{alg:XNAS}.
Thus taking the telescopic product yields,
\begin{align}
\gamma_T = \prod_{t=1}^{T} \Gamma_t 
<
\prod_{t=1}^{T}\frac{|I_{t-1}|}{|I_t|}
= 
\frac{|I_0|}{|I_T|}
=
\frac{N}{|I_T|} 
\leq 
N
\end{align}
Finally we have,
\begin{align}
    1 \leq \gamma_T < N
\end{align}
\qed
\end{proof}

\subsection{Proof of XNAS Regret Bound}
\label{XNAS_regret_proof}
\begin{proof}
\label{XNAS_optimality_proof}
First let us state an auxiliary lemma,
\begin{lemma}[Hoeffding] 
\label{hoeffding}
For any random variable $X$ with $\Pr{\paren{a\leq X \leq b}}=1$ and $s\in \mathbb{R}$, 
\begin{eqnarray}
  \ln{\Exp{e^{sX}}} \leq s\Exp{X} + \frac{s^2(b-a)^2}{8}.
\end{eqnarray}
\end{lemma}
Proof can be found in \cite{hoeffding1953lower}.

We start with defining experts' auxiliary and accumulated-auxiliary losses following \cite{cesa2006prediction} ,
\begin{eqnarray}
\ell'_t(f_{t,i})&=& \nabla\ell (p_t)\cdot f_{t,i}= -R_{t,i} \\
L'_{T,i}&=&\sum_{t=1}^T \ell'_t(f_{t,i})
\end{eqnarray}
Notice that the auxiliary losses are bounded by an input parameter in line \ref{grad_bound} of algorithm~\ref{alg:XNAS}, 
\begin{align}
|\ell'_t(f_{t,i})| \leq \mathcal{L}
\end{align}
We also define the set of remaining experts at time $t$,
\begin{align}
    I_t=\braces{i|i \in I_{t-1} \text{ and } v_{t,i} \geq \theta_t} \quad ; \quad I_0 = \{1, \dots, N\}
\end{align}
Such that,
\begin{align} \label{eq:subset}
    I_T \subset I_{T-1} \subset \dots \subset I_0
\end{align}
We Now bound the ratio of weights sums from both sides. \\
Let us derive a lower bound:
\begin{align}
\ln{\frac{V_T}{V_0}}&=\ln{\frac{\sum_{i\in I_T} v_{T,i}}{N}} \notag\\
&= \ln\paren{{\sum_{i\in I_T}\exp{\{-\eta L'_{T,i}}\}}} -\ln{N} \notag\\
&\geq \ln{ \max_{i\in I_T} \exp{\{-\eta L'_{T,i}}\}} -\ln{N} \notag  \\
&= \ln{ \max_{i=1,\dots,N} \exp{\{-\eta L'_{T,i}}\}} -\ln{N} \label{sum_to_max}  \\
&=-\eta \min_{i=1,\dots,N}{L'_{T,i}} -\ln{N} \notag 
\end{align}

Where we used lemma \ref{no_wipe} in \eqref{sum_to_max}, assuring that loss minimizer is among the remaining experts.\\
Let us derive an upper bound:
\begin{align}
\ln{\frac{V_t}{V_{t-1}}} 
&=
\ln{\frac{\sum_{i\in I_t}  v_{t,i}}{\sum_{i \in I_{t-1}}  v_{t-1,i}}} \notag 
\\ &=
\ln{\frac{\sum_{i\in I_{t-1}}  v_{t,i}}{\sum_{i \in I_{t-1}}  v_{t-1,i}}} + \ln{\frac{\sum_{i\in I_t}  v_{t,i}}{\sum_{i \in I_{t-1}} v_{t,i}}}\notag 
\\ &=
\ln{\frac{\sum_{i\in I_{t-1}}  v_{t,i}}{\sum_{i \in I_{t-1}}  v_{t-1,i}}} + \ln{\frac{\sum_{i\in I_t}  v_{t,i}}{\sum_{i \in I_t} v_{t,i} + \sum_{i \in I_{t-1}\setminus I_t} v_{t,i}}}\label{eq:due_to_subset}
\\ &=
\ln{\frac{\sum_{i\in I_{t-1}}  v_{t,i}}{\sum_{i \in I_{t-1}}  v_{t-1,i}}} - \ln{\Gamma_t}\label{eq:declare_gamma_t}
\\ &= 
\ln{\frac{\sum_{i\in I_{t-1}}  v_{t-1,i} \cdot \exp{\{-\eta \ell'_t(f_{t,i})\}}} {\sum_{i \in I_{t-1}}  v_{t-1,i}}} - \ln{\Gamma_t} \notag
\\ &\leq 
-\eta \frac{\sum_{i\in I_t} v_{t-1,i} \cdot \ell'_t(f_{t,i})}{\sum_{i\in I_t}  v_{t-1,i}} + \frac{\eta^2 \mathcal{L}^2}{2} - \ln{\Gamma_t} \label{from_hoef}
\\ &=
-\eta \ell'_t(p_t)  + \frac{\eta^2 \mathcal{L}^2}{2} - \ln{\Gamma_t} \label{aux_lin}
\end{align}
Where \eqref{eq:due_to_subset} is due to \eqref{eq:subset} and 
\eqref{eq:declare_gamma_t} is by setting, 
\begin{align}
    \Gamma_t := 1 + \frac{\sum_{i \in I_{t-1}\setminus I_t} v_{t,i}}{\sum_{i\in I_t}  v_{t,i}} 
    =
    \frac{\sum_{i \in I_t} v_{t,i} + \sum_{i \in I_{t-1}\setminus I_t} v_{t,i}}{\sum_{i\in I_t}  v_{t,i}}
\end{align}
Inequality \eqref{from_hoef} results from lemma \ref{hoeffding}, with $[a,b]=[-\mathcal{L},\mathcal{L}]$ and \eqref{aux_lin} is a result of the linearity of $\ell'_t$.

Summing the logs telescopic sum:
\begin{align}
 \ln{\frac{V_T}{V_0}} = \sum_{t=1}^T\ln{\frac{V_t}{V_{t-1}}} \leq -\eta \sum_{t=1}^T \ell'_t(p_t)  + \frac{\eta^2 T \mathcal{L}^2}{2} - \sum_{t=1}^T \ln{\Gamma_t}
\end{align}

Setting $\gamma_T :=\prod_{t=1}^T\Gamma_t$ with bounds specified in by lemma \ref{lemma:gamma_bound}, we have,
\begin{align}
 \ln{\frac{V_T}{V_0}} \leq -\eta \sum_{t=1}^T \ell'_t(p_t)  + \frac{\eta^2 T \mathcal{L}^2}{2} - \ln{\gamma_T}
\end{align}
Combining the lower and upper bounds and dividing by $\eta$,
\begin{align}
\label{ell_bound}
\sum_{t=1}^T  \ell'_t(p_t)- \min_{i=1,\dots,N}{L'_{T,i}}  \leq \frac{\eta T\mathcal{L}^2}{2} +\frac{1}{\eta}\ln{N} -\frac{1}{\eta}\ln{\gamma_T}
\end{align}

We now bound the accumulated regret, using the convexity of the loss,
\begin{eqnarray}
\mathrm{Regret}_T&=&\sum_{t=1}^T \ell_t(p_t) - \min_{i=1,\dots,N} \sum_{t=1}^T \ell_t({f_{t,i})} \notag \\
&=& \max_{i=1,\dots,N} \braces{\sum_{t=1}^T \ell_t (p_t) -\ell_t {(f_{t,i})}} \notag \notag \\
&\leq& \max_{i=1,\dots,N} \braces{\sum_{t=1}^T \nabla\ell_t(p_t) \cdot \paren{p_t-f_{t,i}}} \notag \\
&=& \max_{i=1,\dots,N} \braces{ \sum_{t=1}^T \ell'_t(p_t)-\ell'_t(f_{t,i})} \notag \\
&=& \sum_{t=1}^T  \ell'_t(p_t)- \min_{i=1,\dots,N}{L'_{T,i}}
\label{regret_ell_bound}
\end{eqnarray}

Combining \ref{ell_bound} and \ref{regret_ell_bound} completes the proof,
\begin{eqnarray}
\mathrm{Regret}_T \leq \frac{\eta T \mathcal{L}^2}{2} +\frac{1}{\eta}\ln N -\frac{1}{\eta}\ln{\gamma_T}
\end{eqnarray}
\qed
\end{proof}

\section{Supporting Materials for the Key Properties Discussion}
    \label{sec:key_properties_supports}
    \subsection{A Deterministic 3D Axes Toy Problem}
    \label{sec:deterministic_toy}
In an attempt to demonstrate the possible recovery of \emph{late bloomers}, we view an optimization problem in a three dimensional space as a prediction-with-experts problem, where each axis represents an expert with a constant prediction, i.e. $f_{t,x}\equiv(1,0,0), f_{t,y}\equiv(0,1,0), f_{t,z}\equiv(0,0,1)$ for the $x,y,z$ axes respectively. The forecaster then makes a prediction according to the following,
\begin{align*}
    p_t = \frac{v_{t,x}\cdot f_{t,x} + v_{t,y}\cdot f_{t,y} + v_{t,z}\cdot f_{t,z}}{v_{t,x}+v_{t,y}+v_{t,z}} = x_t\cdot f_{t,x} + y_t\cdot f_{t,y} + z_t\cdot f_{t,z} = (x_t,y_t,z_t)
\end{align*}
with $p_t$ in the simplex, i.e. $(x_t,y_t,z_t)\in\Delta=\left\{(x,y,z)\mid x,y,z \geq 0, x+y+z = 1\right\}$. Setting $v_{t,i}=\exp\{\alpha_{t,i}\}$ for $i\in\{x,y,z\}$, for a given loss function $\ell(p_t)$, the update terms for GD with softmax (\autoref{sec:recovery_of_late_bloomers}) and XNAS (\autoref{sec:arch_search}), associated with the $z$ axis, are as follows,
\begin{align*}
 \text{(\textbf{GD}) } \partial_{\alpha_z} \ell(p_t) =  z ((x+y)\partial_z \ell(p_t) - x\partial_x \ell(p_t) - y\partial_y \ell(p_t)) 
 \,\,\, \text{(\textbf{XNAS}) }\nabla_{p_t}\ell (p_t) f_{t,z} =  \partial_z \ell(p_t)
\end{align*}
as the corresponding terms for the $x$ and $y$ axes are similar by symmetry, see a full derivation in section \ref{sec:derivatives_of_3d_toy}. 

\begin{figure}[ht]
    \begin{subfigure}{}
        \includegraphics[width=0.3\linewidth,scale=1]{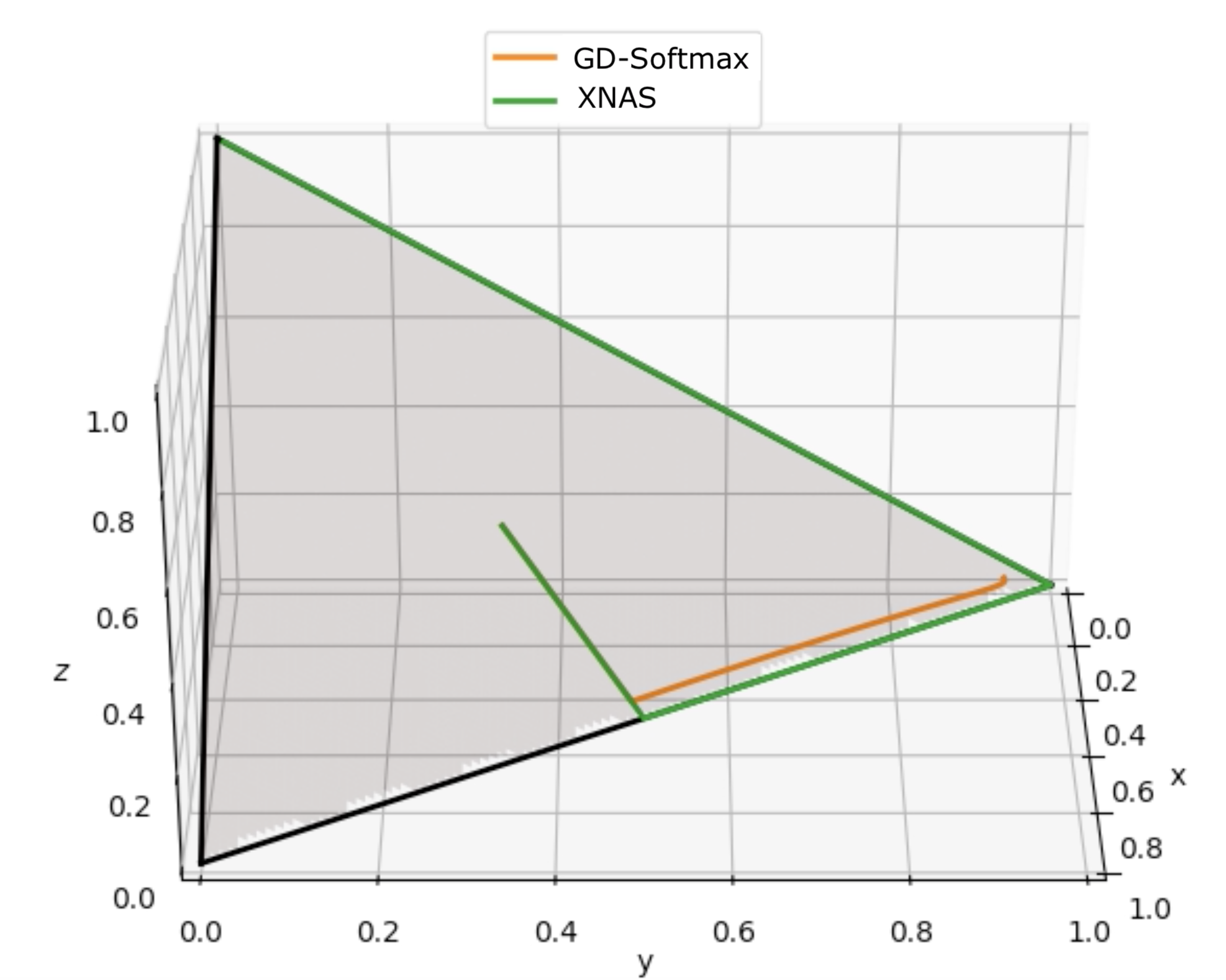} 
    \end{subfigure}
    \begin{subfigure}{}
        \includegraphics[width=0.3\linewidth, ,scale=1]{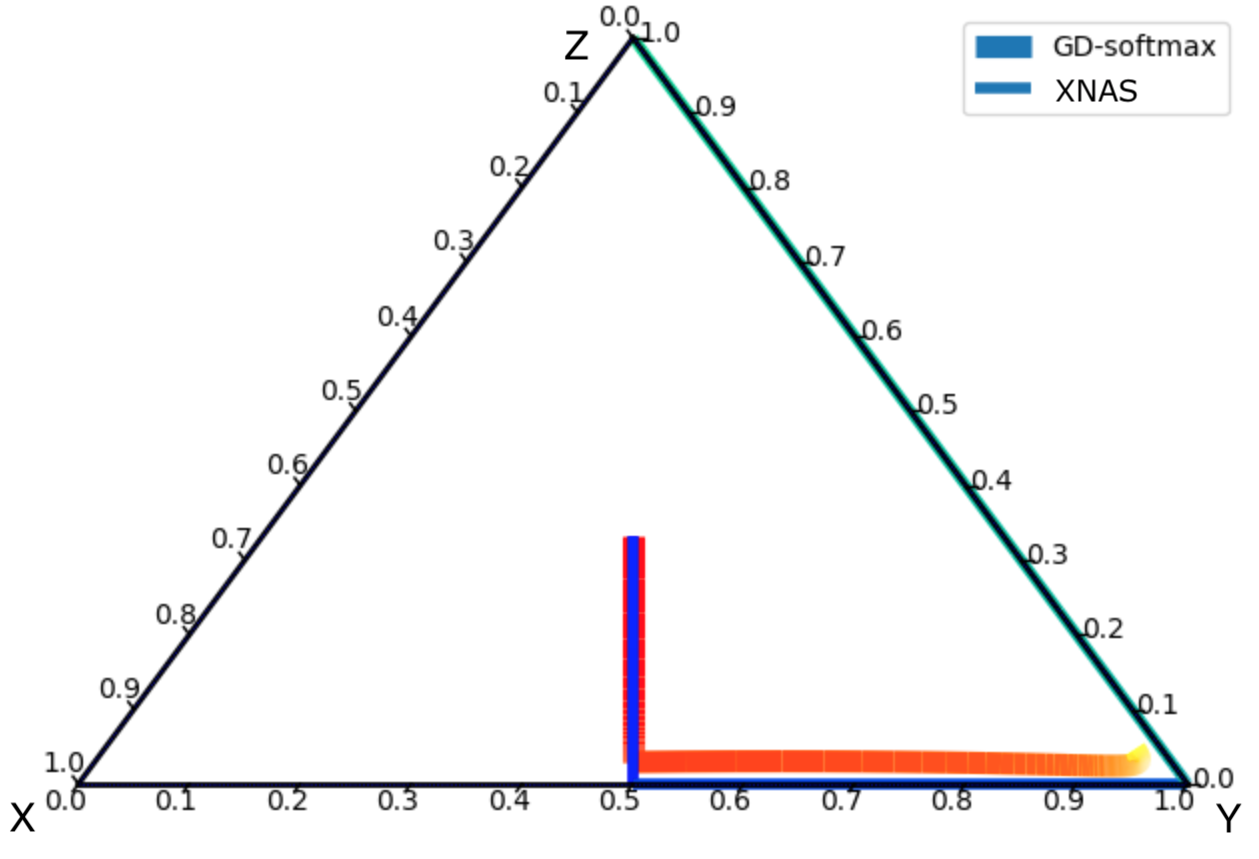}
    \end{subfigure}
    \begin{subfigure}{}
        \includegraphics[width=0.3\linewidth, ,scale=1]{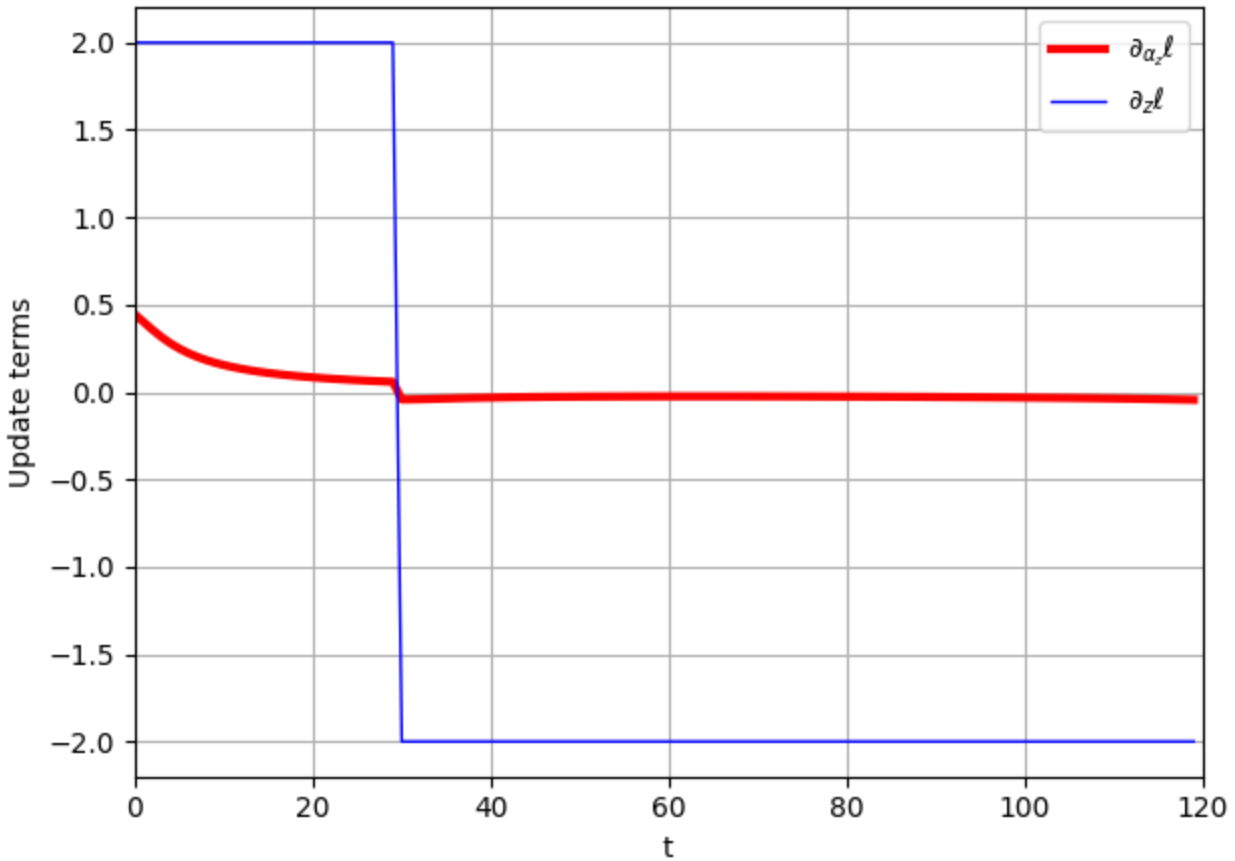}
    \end{subfigure}
    \caption{Trajectories of XNAS and GD with softmax in 3D (left), projected on the simplex (middle) and the update terms for the $z$ axis (right). At the middle, color indicates
progression through time, starting darker and ending lighter}
    \label{fig:xy_deterministic_toy}
\end{figure}

Now let us present the following toy problem: A three dimensional linear loss function $\ell_1(x,y,z) = 2z$ is being optimized for $30$ gradient steps over the simplex, i.e. $(x,y,z)\in\Delta$. Then for illustrating a region shift in terms of the best expert, the loss function shifts into another linear loss function $\ell_2(x,y,z) = -y-2z$ which is then optimized for additional $90$ steps. The trajectories are presented in Figure \ref{fig:xy_deterministic_toy}. 
At the first stage the $z$ axis suffers many penalties as its weight shrinks. Then it starts receiving rewards. For GD with softmax, those rewards are attenuated, as explained in section \ref{sec:recovery_of_late_bloomers} and can be seen in Figure \ref{fig:xy_deterministic_toy} (right). Despite the linear loss, with constant gradients, the update term decays. Note that this effect is even more severe when dealing with more common losses of higher curvature, where the gradients decay near the local minimum and then further attenuated, as illustrated in section \ref{sec:2d_higher_curvature}. Once the gradient shifts, it is already depressed due to past penalties, hence the $z$ axis struggles to recover. XNAS, however, is agnostic to the order of the loss values in time. Once the rewards balance out the penalties, the path leads towards the $z$ axis. In the meanwhile, the $y$ axis takes the lead.
    
    \subsection{A Deterministic 2D Axes Toy Problem}
    \label{sec:2d_toy}
In section \ref{sec:deterministic_toy} we show the built-in attenuation of weak operations by GD with softmax. This is illustrated by a three dimensional toy example where the axes represent experts of constant predictions. Here we elaborate on this effect using a similar two dimensional toy problem, where the gradients with respect to the axes are the negative values of one another.
See section \ref{sec:derivative_of_2d_toy} for the setup and full derivations.
All the experiments in this section are conducted using a learning rate of $0.1$ for both optimizers for $50$ steps.

\subsubsection{Balanced Initialization}
\label{sec:balanced_init}
\begin{figure}[ht]
    \begin{subfigure}{}
        \includegraphics[width=0.3\linewidth, height=3cm]{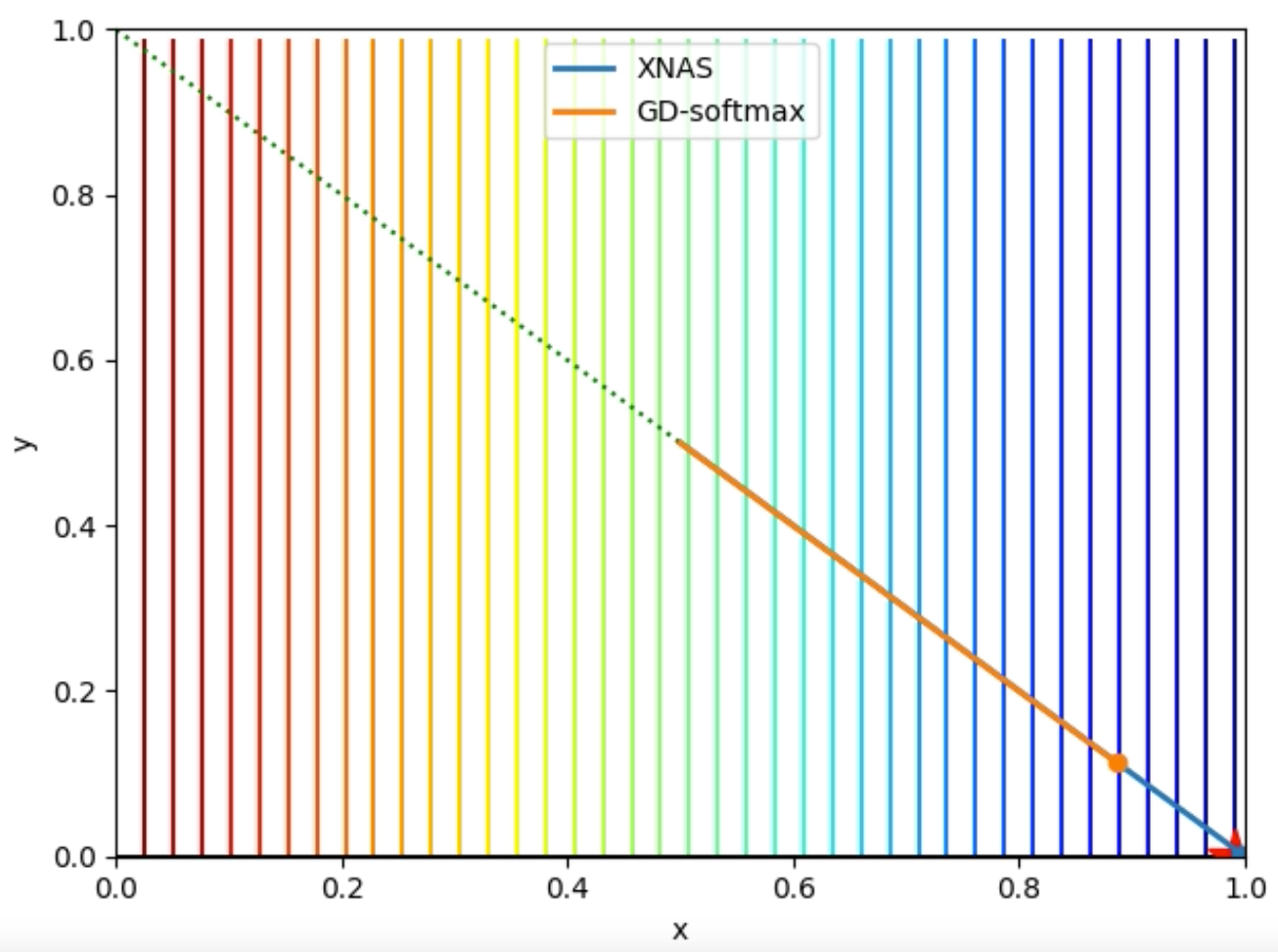} 
    \end{subfigure}
    \begin{subfigure}{}
        \includegraphics[width=0.3\linewidth, height=3cm]{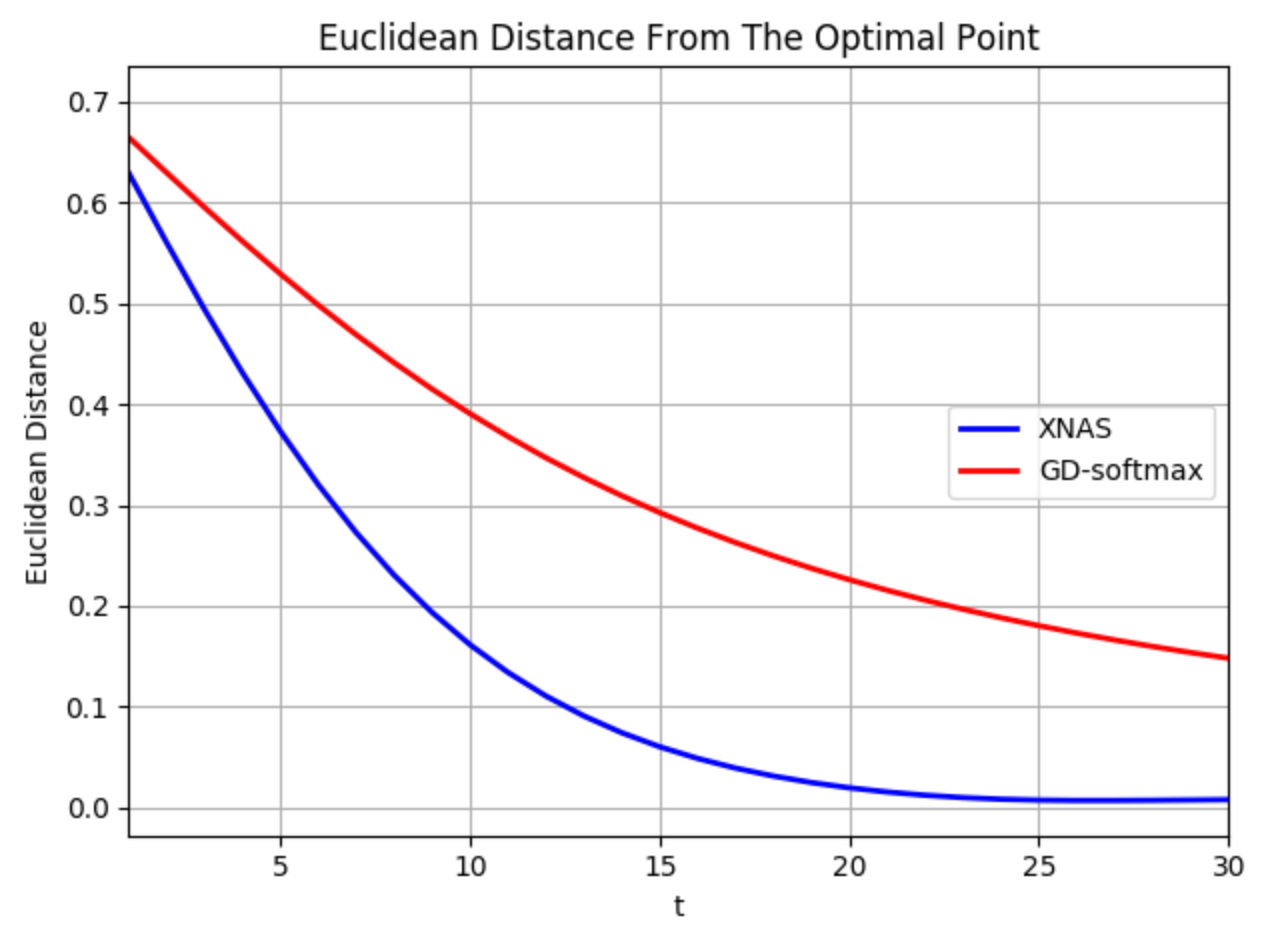}
    \end{subfigure}
    \begin{subfigure}{}
        \includegraphics[width=0.3\linewidth, height=3cm]{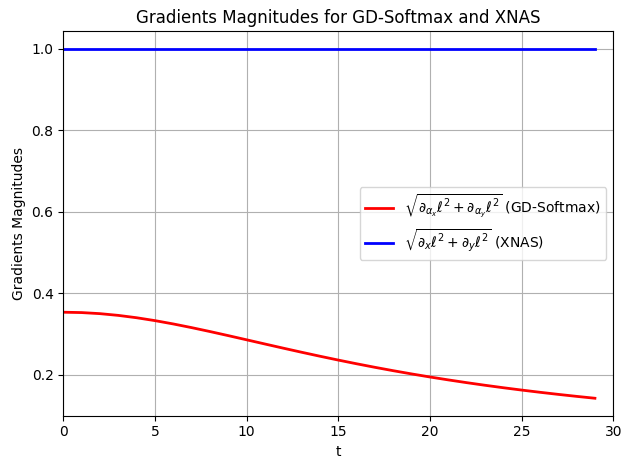}
    \end{subfigure}
    \caption{Trajectories of GD with softmax and XNAS for a $\ell(x,y) = -x$ (left) the distance from the optimal point (middle) and the corresponding magnitude of the gradients (right) for a balanced initialization $\alpha_y = \alpha_x$. The vertical lines are the contour lines of the loss function. The dashed line is the simplex $x+y=1$, and the trajectories are the solid lines with circles at their ends.}
    \label{fig:2d_linear_loss}
\end{figure}
Figure \ref{fig:2d_linear_loss} illustrates the attenuation of gradients for GD with softmax, as although the gradients of the loss are constant, the gradients' magnitude decreases as we move away from the initialization $\alpha_x = \alpha_y$, i.e. $(x,y)=(0.5,0.5)$. XNAS indeed receives constant gradients thus reaches the minima faster. 

\subsubsection{Imbalanced Initialization}
\label{sec:imbalanced_init}
\begin{figure}[ht]
    \begin{subfigure}{}
        \includegraphics[width=0.3\linewidth, height=3cm]{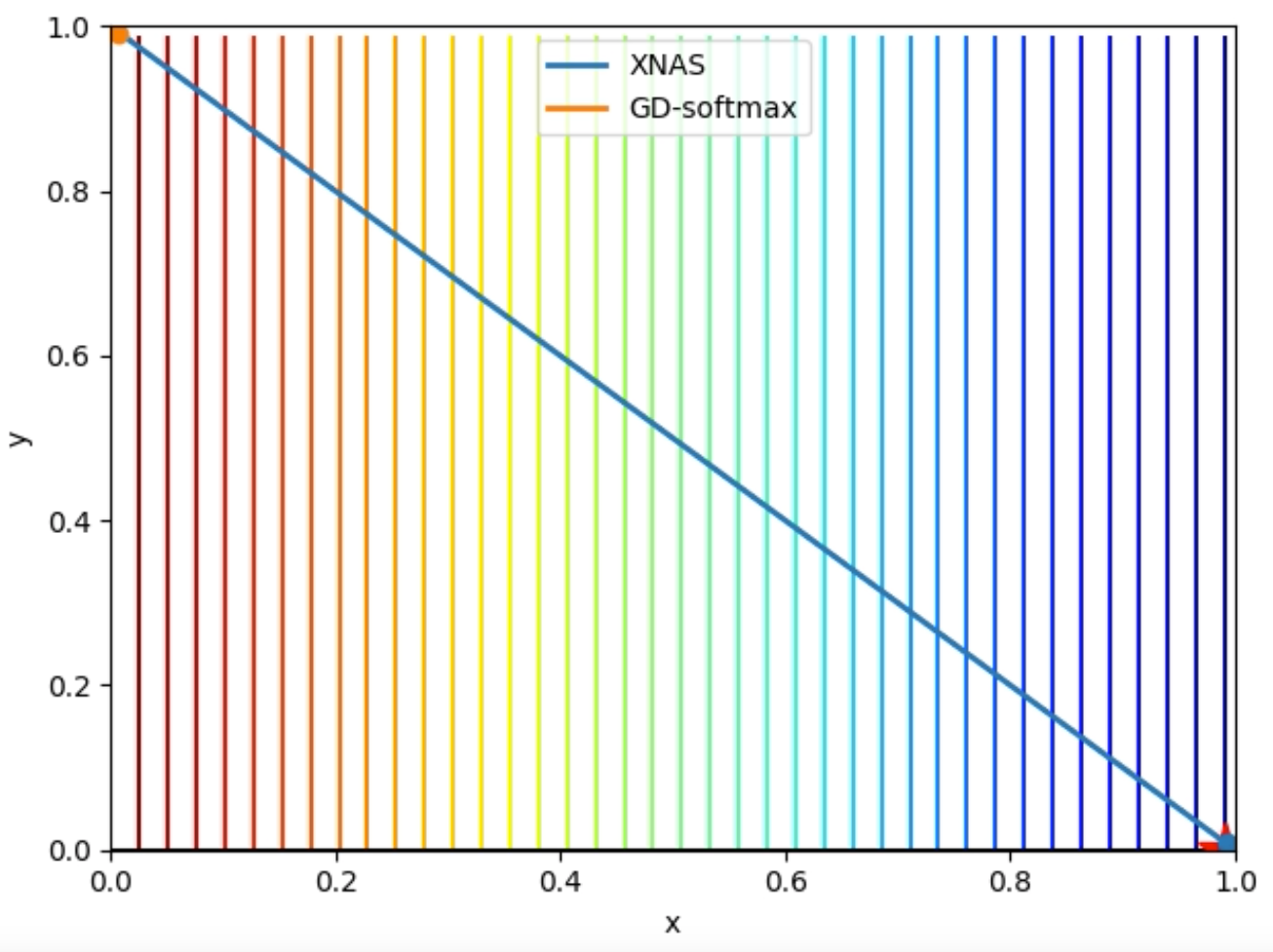} 
    \end{subfigure}
    \begin{subfigure}{}
        \includegraphics[width=0.3\linewidth, height=3cm]{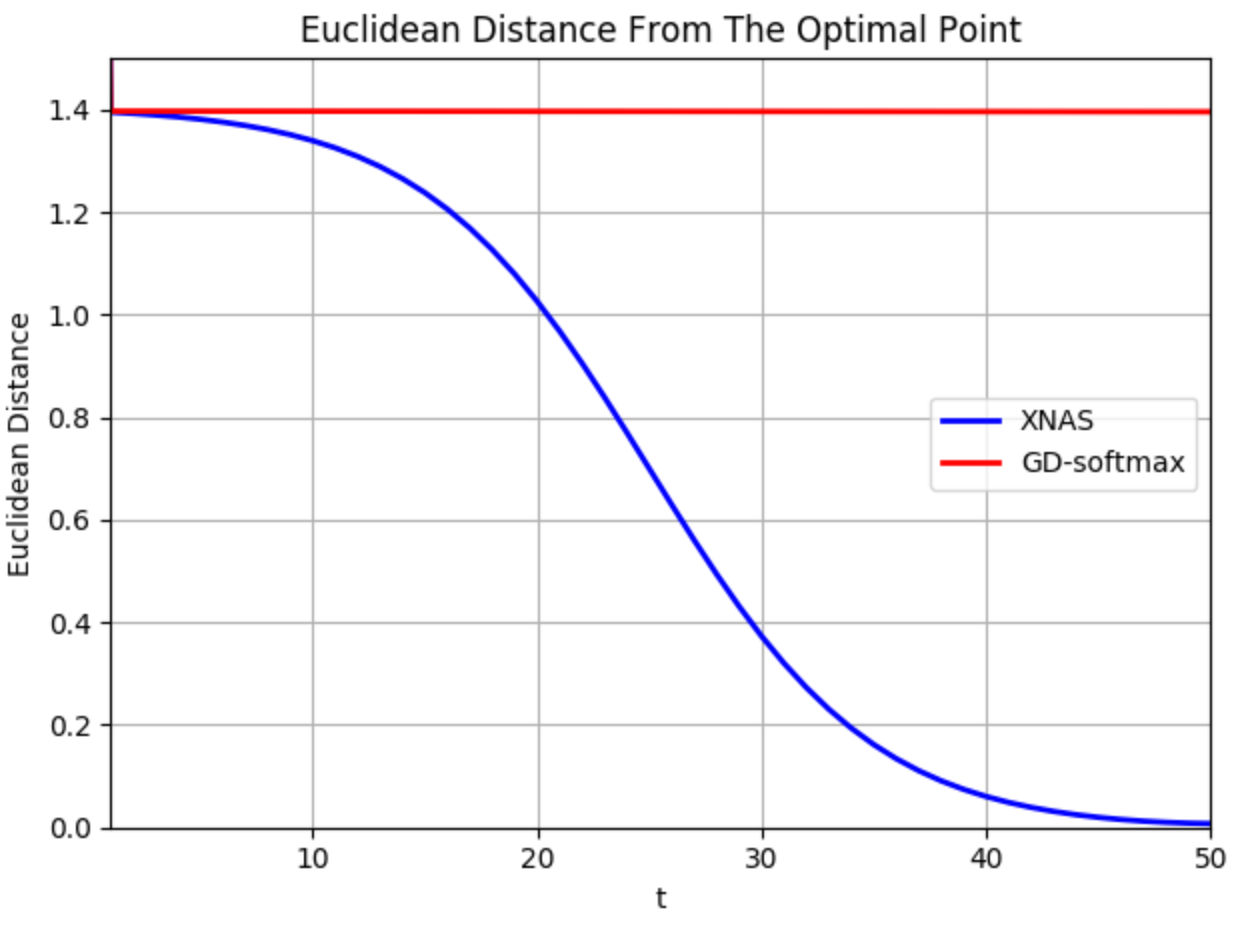}
    \end{subfigure}
    \begin{subfigure}{}
        \includegraphics[width=0.3\linewidth, height=3cm]{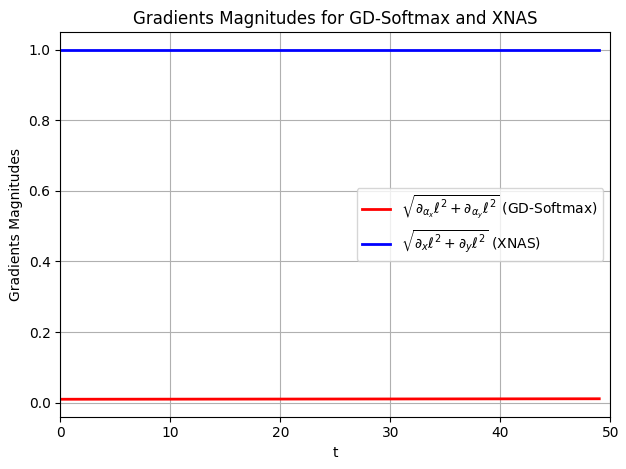}
    \end{subfigure}
    \caption{Trajectories of GD with softmax and XNAS for a $\ell(x,y) = -x$ (left) the distance from the optimal point (middle) and the corresponding magnitude of the gradients (right) for an imbalanced initialization $\alpha_x < \alpha_y$. The vertical lines are the contour lines of the loss function. The dashed line is the simplex $x+y=1$, and the trajectories are the solid lines with circles at their ends.}
    \label{fig:2d_bad_init}
\end{figure}

The attenuated gradients also make GD with softmax more sensitive to the initialization, as demonstrated in Figure \ref{fig:2d_bad_init}, where $\alpha_x = 0 < 5 = \alpha_y$ and GD with softmax, whose gradients are attenuated, makes no progress while XNAS reaches the minima.


\subsubsection{The Preference of Dense Solutions}
\label{sec:the_preference_of_dense_solutions}
\begin{figure}[ht]
    \begin{subfigure}{}
        \includegraphics[width=0.45\linewidth, height=5cm]{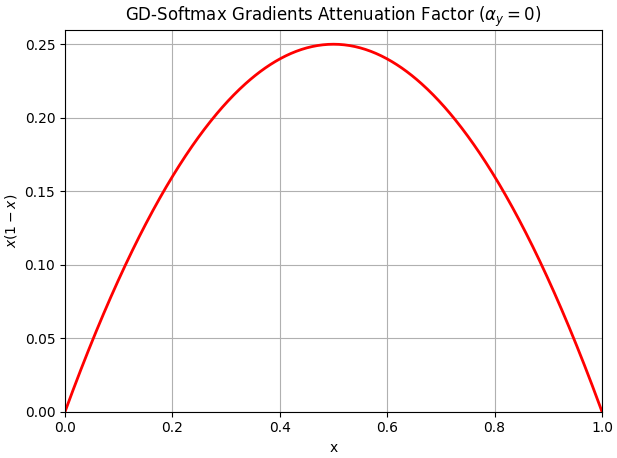}
    \end{subfigure}
    \begin{subfigure}{}
        \includegraphics[width=0.45\linewidth, height=5cm]{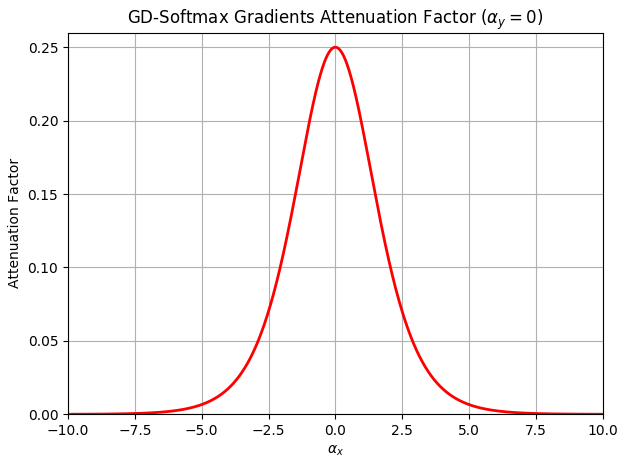}
    \end{subfigure}
    \caption{The attenuation factor of GD with softmax along the simplex (left) and versus $\alpha_x$ as $\alpha_y = 0$ (right)}
    \label{fig:2d_attenuation_factor}
\end{figure}

Presenting the attenuation factor $x\cdot y$ on the simplex, i.e. $x \cdot (1-x)$, in Figure \ref{fig:2d_attenuation_factor}, demonstrates how gradients are harder attenuated as far away as the variables move from a dense solution, e.g. $(x,y)=(0.5,0.5)$ at $\alpha_x = \alpha_y$. 
Hence, it is harder for GD with softmax to strengthen a single expert over the other. This effect encourages dense solutions over sparse solutions, i.e. a choice of a single expert. 
Due to the \emph{descretization stage}, described in section \ref{sec:nueral_architecture_space}, the denser the solution is, the more degradation in performance is incurred. Hence dense solutions should be discouraged rather than encouraged.

\subsubsection{Loss Functions of a Higher Curvature}
\label{sec:2d_higher_curvature}
\begin{figure}[ht]
    \begin{subfigure}{}
        \includegraphics[width=0.3\linewidth, height=3cm]{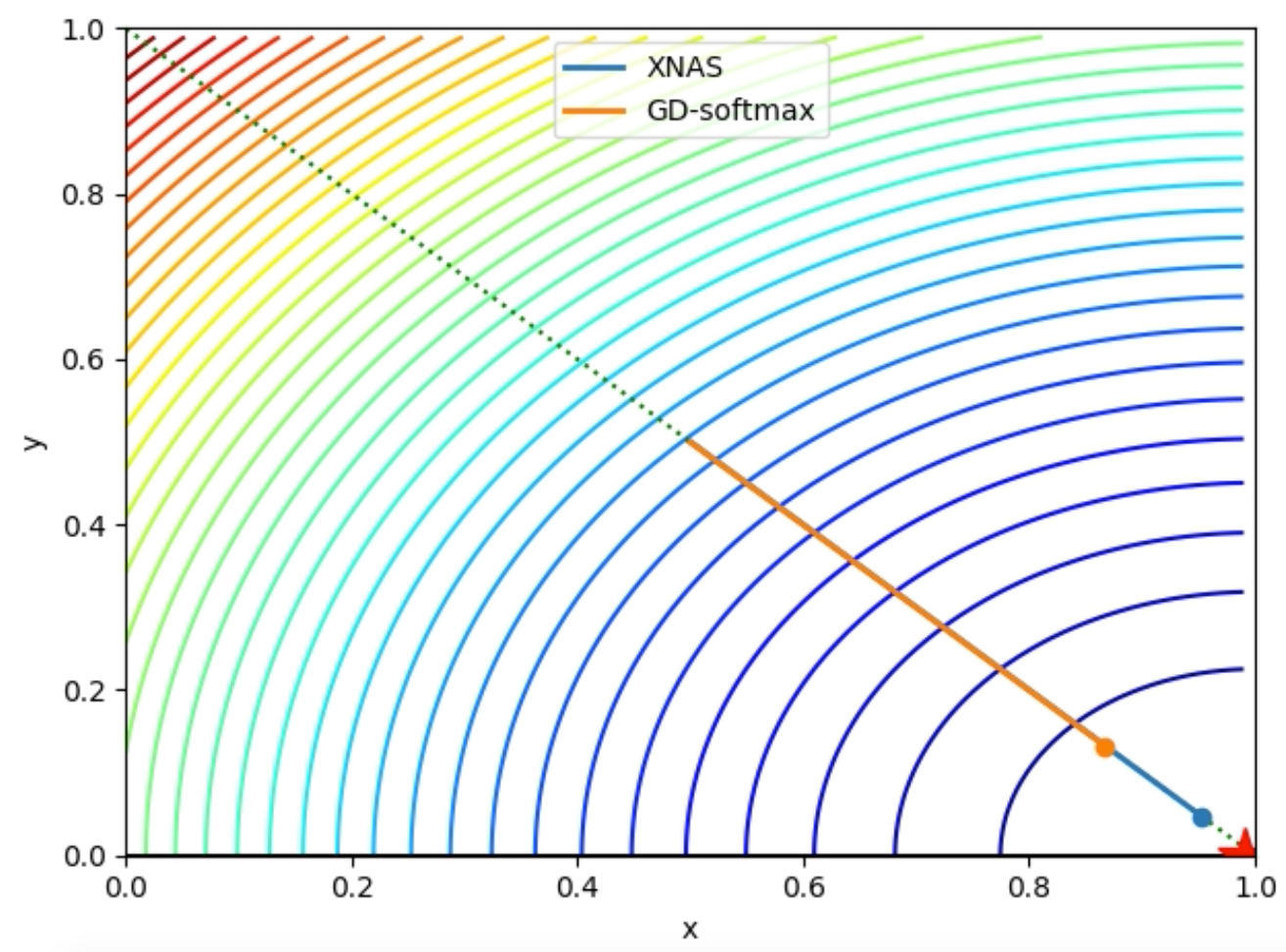} 
    \end{subfigure}
    \begin{subfigure}{}
        \includegraphics[width=0.3\linewidth, height=3cm]{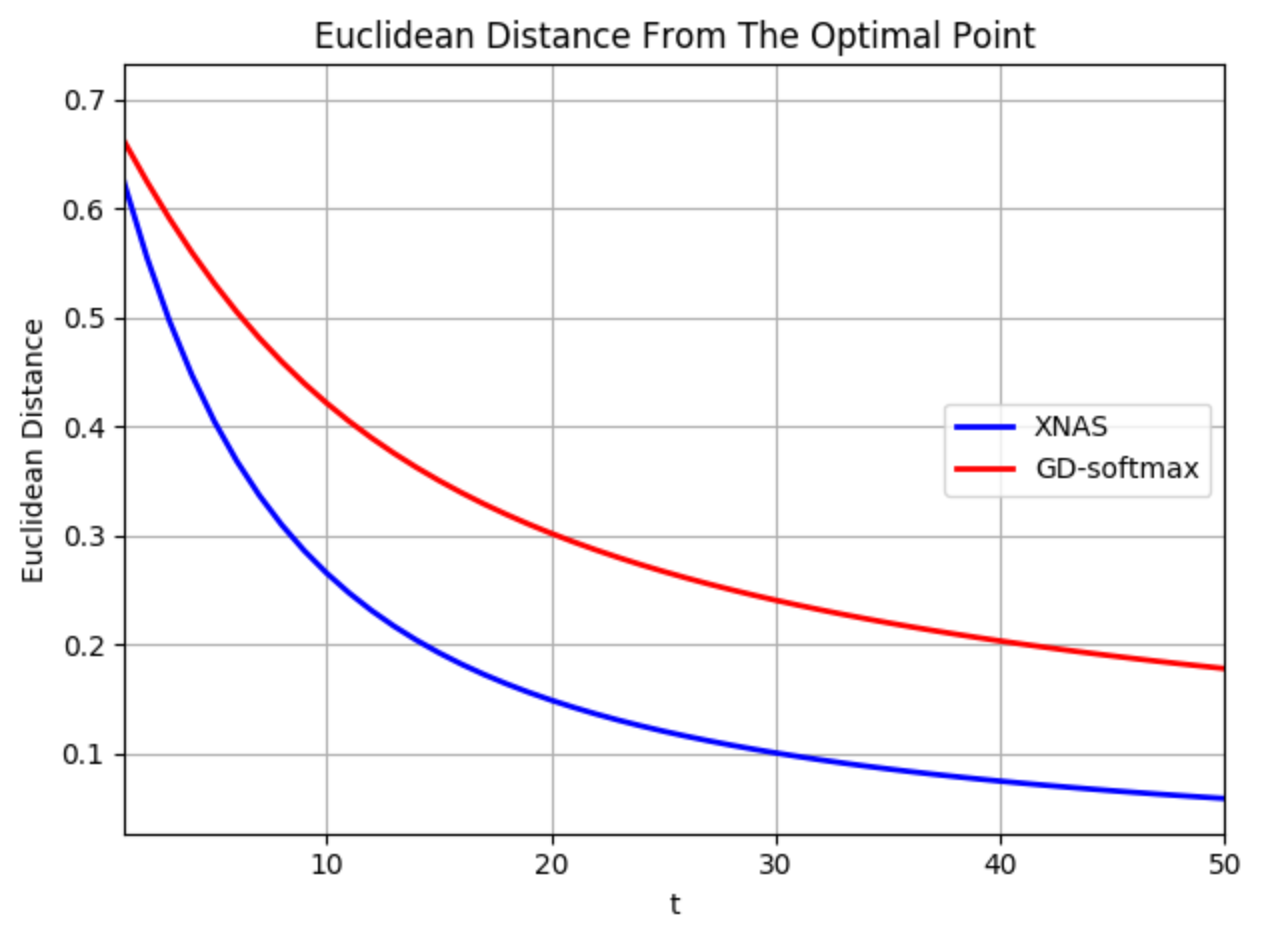}
    \end{subfigure}
    \begin{subfigure}{}
        \includegraphics[width=0.3\linewidth, height=3cm]{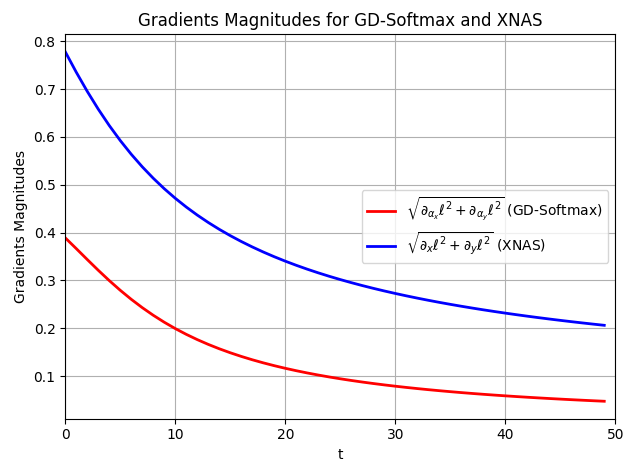}
    \end{subfigure}
    \caption{Trajectories of GD with softmax and XNAS for a $\ell(x,y) = 0.5(x-1)^2 + 0.5y^2$ (left)  the distance from the optimal point (middle) and the corresponding magnitude of the gradients (right). The circular lines are the contour lines of the loss function. The dashed line is the simplex $x+y=1$, and the trajectories are the solid lines with circles at their ends.}
    \label{fig:2d_quadratic}
\end{figure}

Sections \ref{sec:balanced_init} and \ref{sec:imbalanced_init} deal with a linear loss function of no curvature and constant gradients. Once convex loss functions are considered, the gradients decrease towards the local minimum. Figure \ref{fig:2d_quadratic} illustrates the further attenuation of GD with softmax for a quadratic loss function $\ell(x,y) = 0.5(x-1)^2 + 0.5y^2$, which makes it even harder for GD with softmax to reach the local minimum.

    \subsection{Regret and Correct Selection in Statistical Setting}
    \label{sec:stochastic_toy}
\begin{figure}[ht]
 \label{fig:sto_toys}
    \begin{subfigure}{}
        \includegraphics[width=0.45\linewidth, height=5cm]{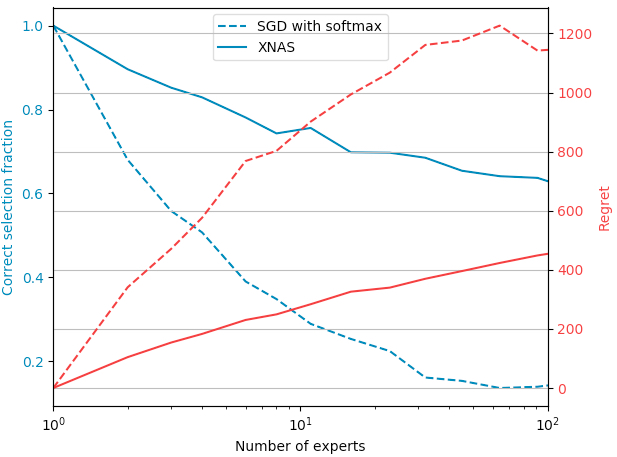} 
        \label{fig:sto1}
    \end{subfigure}
    \begin{subfigure}{}
        \includegraphics[width=0.45\linewidth, height=5cm]{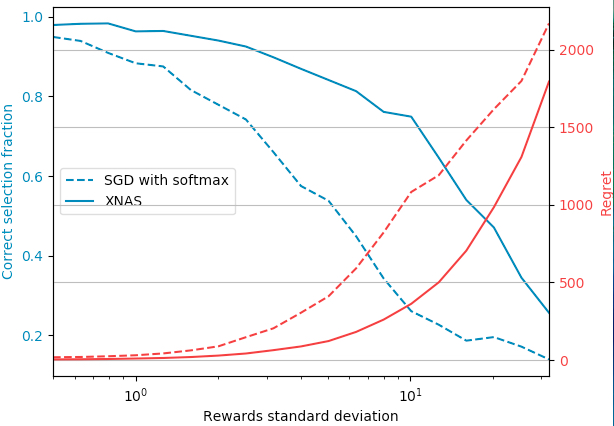}
        \label{fig:sto2}
    \end{subfigure}
    \caption{Correct selection fraction and regret of optimizers.}
    \label{fig:sto}
\end{figure}

We consider a statistical setup for comparing XNAS with the common Gradient Descent (GD) with softmax, described in section \ref{sec:recovery_of_late_bloomers}. This setup simulates the iterative architecture optimization and final selection of the top expert for a single forecaster. 

Two forecasters are compared, XNAS and GD with softmax. Both receive noisy independent and identically distributed (i.i.d) rewards of $N$ experts.
Each expert has an initial i.i.d bias $\left\{b_i\right\}_{i=1}^N\sim \textrm{N}(0,1)$ simulating its inherent value, so its rewards satisfy $R_{t,i}\sim \textrm{N}(b_i,\sigma_R^2)$ for $i=1,\dots,N$ and $t=1,\dots,T$, where $\textrm{N}(\mu,\sigma^2)$ is a Gaussian distribution with a mean $\mu$ and a standard deviation $\sigma$.

The first forecaster updates its weights using GD with softmax update rule from ~\autoref{GD_SM_update} (full derivation in \autoref{GD_SM_derivation}), common to previous NAS methods, while the second is using Algorithm~\ref{alg:XNAS}.

The forecasters use their update rules to update weights along the run. At the end of the run each selects the expert with the largest weight. A correct classification satisfies $\max\limits_{i=1,\dots,N}{\alpha_i}=\max\limits_{i=1,\dots,N}{b_i}$. The average regret of those runs is also calculated based on equation~\ref{losses_and_regret}.

Figure \ref{fig:sto} shows a mean of $1000$ Monte-Carlo runs, each of $1000$ time-steps, plotting the regret and the fraction of correct selection (classification).
In Figure \ref{fig:sto} (left), both terms are plotted versus a varying number of experts. It can be seen that the regret of XNAS is significantly smaller, scaling with the number of experts like $O(\sqrt{\ln{N}})$, as implied by its regret upper bound in equation~\ref{eq:eta_and_regret}, while GD regret scales like $O(\sqrt{N})$~\cite{hazan2016introduction}. 

In Figure \ref{fig:sto} (right), the noise standard deviation $\sigma_R$ is varying, making it harder to correctly classify the expert with the highest bias. Again, XNAS dominates GD with softmax, which is more sensitive to the noisy rewards due to the 'late bloomers' described in \ref{sec:recovery_of_late_bloomers}, e.g. the best experts might suffer some large penalties right at the beginning due to the noise, thus might not recover for GD with softmax. \\
In both graphs it can be seen that the correct selection fraction is monotonically decreasing as the regret is increasing. This gives an additional motivation for the use of the regret minimization approach as a criterion for neural architecture search.
    
    \subsection{Gradients Derivations}
    \label{sec:gradients_derivations}

For the comparison to previous work \cite{liu2018darts}, we consider the decision variables $\alpha_{t,i} = \ln{v_{t,i}}$, as the right hand side is defined at \eqref{eq:forecaster}.
\subsubsection{The Derivation of Derivatives in the General Case} \label{GD_SM_derivation}
\begin{align}
    \frac{\partial u_{t,j}}{\partial\alpha_{t,i}} 
    &=
    \frac{\partial}{\partial\alpha_{t,i}} \frac{e^{\alpha_{t,j}}}{\sum_{k=1}^N e^{\alpha_{t,k}}}
    \notag\\&= 
    \frac{\delta_{i,j}e^{\alpha_{t,i}}\cdot \sum_{k=1}^N e^{\alpha_{t,k}} - e^{\alpha_{t,j}} \cdot e^{\alpha_{t,i}}}{\paren{\sum_{k=1}^N e^{\alpha_{t,k}}}^2}
    \notag\\&= 
    \frac{e^{\alpha_{t,i}}}{\sum_{k=1}^N e^{\alpha_{t,k}}}\paren{\delta_{i,j} - \frac{e^{\alpha_{t,j}} }{\sum_{k=1}^N e^{\alpha_{t,k}}}}
    \notag\\&= 
     u_{t,i}\paren{\delta_{i,j} -  u_{t,j}} 
    \label{delta_derivative}
\end{align}

where $\delta_{i,j} =
    \begin{cases}
            1, &         \text{if } i=j,\\
            0, &         \text{if } i\neq j.
    \end{cases}$ is the Kronecker delta.
    
Observe that:
\begin{align}
    \frac{\partial\ell_t(p_t)}{\partial u_{t,i}} 
    &=
    \nabla_{  p_t}\ell_t(  p_t)^T\cdot 
    \frac{\partial  p_t}{\partial{ u_{t,i}}}
    \notag\\&=
    \nabla_{  p_t}\ell_t(  p_t)^T\cdot 
    \frac{\partial}{\partial{ u_{t,i}}}\sum_{j=1}^N  u_{t,j}   f_{t,j}
    \notag\\&=
    \nabla_{  p_t}\ell_t(  p_t)^T\cdot   f_{t,i}
    \label{eq:derivative_wrt_w_bar}
\end{align}

Finally,
\begin{align}
    \frac{\partial\ell_t(p_t)}{\partial\alpha_{t,i}} 
    &=
    \sum_{j=1}^N \frac{\partial\ell_t(p_t)}{\partial u_{t,j}} \cdot \frac{\partial u_{t,j}}{\partial\alpha_{t,i}}
    \notag\\&=
    \sum_{j=1}^N \nabla_{  p_t}\ell_t(  p_t)^T\cdot   f_{t,j} \cdot   u_{t,i}\paren{\delta_{i,j} -  u_{t,j}}
    \label{eq:setting_prev}\\&=
    \nabla_{  p_t}\ell_t(  p_t)^T \cdot  u_{t,i}\paren{  f_{t,i} - \sum_{j=1}^N  u_{t,j}   f_{t,j}}
    \notag\\&=
    \nabla_{  p_t}\ell_t(  p_t)^T \cdot  u_{t,i}\paren{  f_{t,i} -   p_t}
    \label{eq:derivative_wrt_alpha}
\end{align}
where \eqref{eq:setting_prev} is due to \eqref{delta_derivative} and \eqref{eq:derivative_wrt_w_bar}.

\subsubsection{The Derivation of Derivatives for the 3D Axes Problem}
\label{sec:derivatives_of_3d_toy}
In this section we derive the derivatives with respect to the $x,y,z$ axes for the toy problem introduced at section \ref{sec:deterministic_toy}. In this case, 
\begin{align}
      f_{t,x} &\equiv (1,0,0)^T    &     u_{t,x} &= x_t \notag\\
      f_{t,y} &\equiv (0,1,0)^T    &     u_{t,y} &= y_t \label{eq:3d_case}\\
      f_{t,z} &\equiv (0,0,1)^T    &     u_{t,z} &= z_t \notag
\end{align}
Hence,
\begin{align}
   p_t = \sum_{i\in\{x,y,z\}}  u_{t,i}   f_{t,i} = x_t(1,0,0)^T + y_t(0,1,0)^T + z_t(0,0,1)^T = (x_t,y_t,z_t)^T
\end{align}
 such that for a loss function $\ell$,
\begin{align}
\nabla_{  p_t}\ell(  p_t)
= 
\paren{\frac{\partial\ell(  p_t)(  p_t)}{\partial x_t}, \frac{\partial\ell(  p_t)(  p_t)}{\partial y_t}, \frac{\partial\ell(  p_t)(  p_t)}{\partial z_t}}^T
=
\paren{\partial_{x_t}\ell(  p_t), \partial_{y_t}\ell(  p_t), \partial_{z_t}\ell(  p_t)}^T
\label{eq:3d_mixed_op}
\end{align}
Then, setting \eqref{eq:3d_case} and \eqref{eq:3d_mixed_op} in \eqref{eq:derivative_wrt_alpha}, we have,
\begin{align}
   \partial_{\alpha_x} \ell(  p_t) = \frac{\partial \ell(  p_t)}{\partial\alpha_x} &= 
\paren{\partial_{x_t}\ell(  p_t), \partial_{y_t}\ell(  p_t), \partial_{z_t}\ell(  p_t)}^T \cdot x_t \paren{(1,0,0)^T-(x_t,y_t,z_t)^T}
\notag\\ &=
\paren{\partial_{x_t}\ell(  p_t), \partial_{y_t}\ell(  p_t), \partial_{z_t}\ell(  p_t)}^T \cdot x_t \paren{y_t + z_t,-y_t,-z_t}^T
\label{eq:3d_simplex_step}\\&=
x_t\paren{(y_t + z_t) \partial_{\alpha_x} \ell(  p_t) -y_t \partial_{\alpha_y} \ell(  p_t) - z_t \partial_{\alpha_z} \ell(  p_t)}
\end{align}
where \eqref{eq:3d_simplex_step} is since $(x,y,z)\in\Delta$, defined in section \ref{sec:deterministic_toy}.
By symmetry we have,
\begin{align}
   \partial_{\alpha_y} \ell(  p_t)
   &= 
   y_t\paren{(x_t + z_t) \partial_{\alpha_y} \ell(  p_t) -x_t \partial_{\alpha_x} \ell(  p_t) - z_t \partial_{\alpha_z} \ell(  p_t)}
   \\
   \partial_{\alpha_z} \ell(  p_t)
   &= 
   z_t\paren{(x_t + y_t) \partial_{\alpha_z} \ell(  p_t) -x_t \partial_{\alpha_x} \ell(  p_t) - y_t \partial_{\alpha_y} \ell(  p_t)}
\end{align}
The update terms for XNAS are according to \eqref{eq:derivative_wrt_w_bar}, 
\begin{align}
   \nabla_{  p_t} \ell(  p_t)^T \cdot f_{t,x} =\paren{\partial_{x_t}\ell(  p_t), \partial_{y_t}\ell(  p_t), \partial_{z_t}\ell(  p_t)} \cdot (1,0,0)^T = \partial_{x_t}\ell(  p_t)
    \\
   \nabla_{  p_t} \ell(  p_t)^T \cdot f_{t,y} =\paren{\partial_{x_t}\ell(  p_t), \partial_{y_t}\ell(  p_t), \partial_{z_t}\ell(  p_t)} \cdot (0,1,0)^T =\partial_{y_t}\ell(  p_t)
    \\
   \nabla_{  p_t} \ell(  p_t)^T \cdot f_{t,z} = \paren{\partial_{x_t}\ell(  p_t), \partial_{y_t}\ell(  p_t), \partial_{z_t}\ell(  p_t)} \cdot (0,0,1)^T = \partial_{z_t}\ell(  p_t)
\end{align}
\subsubsection{The Derivation of Derivatives for a 2D Axes Problem}
\label{sec:derivative_of_2d_toy}
In this section we derive the derivatives with respect to the $x,y$ axes for a two dimensional toy problem. Similar to section \ref{sec:deterministic_toy} where a three dimensional problem was considered, now we consider only two axes. Each axis represents an expert of a constant prediction,
\begin{align}
      f_{t,x} &\equiv (1,0)^T    &     u_{t,x} &= x_t \notag\\
      f_{t,y} &\equiv (0,1)^T    &     u_{t,y} &= y_t \label{eq:2d_case}
\end{align}
Where $x_t + y_t \equiv 1$. Hence,
\begin{align}
   p_t = \sum_{i\in\{x,y\}}  u_{t,i}   f_{t,i} = x_t(1,0)^T + y_t(0,1)^T = (x_t, y_t)^T
\end{align}
 such that for a loss function $\ell$,
\begin{align}
\nabla_{  p_t}\ell(  p_t)
= 
\paren{\frac{\partial\ell(  p_t)}{\partial x_t}, \frac{\partial\ell(  p_t)}{\partial y_t}}^T
=
\paren{\partial_{x_t}\ell(  p_t), \partial_{y_t}\ell(  p_t)}^T
\label{eq:2d_mixed_op}
\end{align}
Then, setting \eqref{eq:2d_case} and \eqref{eq:2d_mixed_op} in \eqref{eq:derivative_wrt_alpha}, we have,
\begin{align}
   \partial_{\alpha_x} \ell(  p_t) = \frac{\partial \ell(  p_t)}{\partial\alpha_x} &= 
\paren{\partial_{x_t}\ell(  p_t), \partial_{y_t}\ell(  p_t)}^T \cdot x_t \paren{(1,0)^T-(x_t,y_t)^T}
\notag\\ &=
\paren{\partial_{x_t}\ell(  p_t), \partial_{y_t}\ell(  p_t)}^T \cdot x_t \paren{y_t,-y_t}^T
\label{eq:2d_simplex_step}\\&=
x_t y_t \paren{\partial_{x_t}\ell(  p_t) - \partial_{y_t}\ell(  p_t)}
\end{align}
where \eqref{eq:2d_simplex_step} is since $x_t + y_t \equiv 1$.
By symmetry we have,
\begin{align}
   \partial_{\alpha_y} \ell(  p_t)
   &= 
    x_t y_t \paren{\partial_{y_t}\ell(  p_t) - \partial_{x_t}\ell(  p_t)}
    =
    -\partial_{\alpha_x} \ell(  p_t)
\end{align}

    \subsection{The Mean Normalized Entropy}
    \label{sec:mean_normalized_entropy}
In this section we provide the technical calculation details of the mean normalized entropy, referred to in section \ref{sec:no_weight_decay}.
The normalized entropy of forcaster $(i,j)$ is calculated at the end of the search as following, 
\begin{align}
    \bar{H}_T^{(i,j)} =-\frac{1}{\ln{(N)}}\sum_{i=1}^N u^{(i,j)}_{T,i}\ln{\paren{u^{(i,j)}_{T,i}}}
\end{align} 

The mean is taken over all the forecasters in a normal cell, i.e.
\begin{align}
    \bar{H}_T = \frac{1}{|\mathcal{I}|\cdot|\mathcal{J}|} \sum_{i\in\mathcal{I}}\sum_{j\in\mathcal{J}}\bar{H}_T^{(i,j)}
\end{align} 
where $\mathcal{I}$ and $\mathcal{J}$ are the sets of indices $i$ and $j$ respectively.

\section{Detailed Experiments Setting}

\subsection{Classification Datasets Details}

In this section we will describe the additional datasets that were used for transferability tests in section \ref{Transfer learning results}

{\bf {CINIC-10:} \cite{darlow2018cinic}}
 is an extension of CIFAR-$10$ by ImageNet images, down-sampled to match the image size of CIFAR-$10$. 
 It has $270,000$ images of $10$ classes, i.e. it has larger train and test sets than those of CIFAR-$10$. 

{\bf {CIFAR-100:} \cite{cifar100}}
A natural image classification dataset, containing $100$ classes with $600$ images per class. The image size is $32$x$32$ and the train-test split is $50,000$:$10,000$ images respectively. 

{\bf {FREIBURG:} \cite{Freiburg}}
A groceries classification dataset consisting of $5000$ images of size $256$x$256$, divided into $25$ categories.
It has imbalanced class sizes ranging from $97$ to $370$ images per class. Images were taken in various aspect ratios and padded to squares.

{\bf {SVHN:} \cite{SVHN}}
A dataset containing real-world images of digits and numbers in natural scenes. It consists of $600,000$ images of size $32$x$32$, divided into $10$ classes. The dataset can be thought of as a real-world alternative to MNIST, with an order of magnitude more images and significantly harder real-world scenarios. 

{\bf {FMNIST:} \cite{fashionMnist}}
A clothes classification dataset with a $60,000$:$10,000$ train-test split. Each example is a grayscale image of size $28$x$28$, associated with a label from $10$ classes of clothes. It is intended to serve as a direct drop-in replacement for the original MNIST dataset as a benchmark for machine learning algorithms.

\subsection{CIFAR-10 XNAS Search details}
{\bf Data pre-processing.} We apply the following:
\begin{itemize}
    \item Centrally padding the training images to a size of $40$x$40$.
    \item Randomly cropping back to the size of $32$x$32$.
    \item Randomly flipping the training images horizontally.
    \item Auto augment.    
    \item Standardizing the train and validation sets to be of a zero-mean and a unit variance.
\end{itemize}
{\bf Operations and cells.} We select from the operations mentioned in \ref{arch_search_cifar}, used with stride $1$ on normal cells, and with stride $2$ on reduction cells in edges connected to the two previous cells. Other edges in reduction cells are used with stride $1$.
Convolutional layers are padded so that the spatial resolution is kept. The operations are applied in the order of ReLU-Conv-BN. Following \cite{noy2019asap},\cite{Real18Regularized}, depthwise separable convolutions are always applied twice. 
The cell's output is a $1$x$1$ convolutional layer applied on all of the cells' four intermediate nodes' outputs concatenated, such that the number of channels is preserved.
In CIFAR-$10$, the search lasts up to $0.3$ days on NVIDIA GTX 1080Ti GPU.

\subsection{Train Details}
{\bf CIFAR-10.} \label{cifar10_training}
The training architecture consists of stacking up $20$ cells: $18$ normal cells and $2$ reduction cells, located at the $1/3$ and $2/3$ of the total network depth respectively.
For the three architectures XNAS-Small, XNAS-Medium and XNAS-Large, the normal cells start with $36$, $44$ and $50$ channels respectively, where we double the number of channels after each reduction cell.
We trained the network for $1500$ epochs using a batch size of $96$ and SGD optimizer with nesterov-momentum of $0.9$.
Our learning rate regime was composed of $5$ cycles of power cosine annealing learning rate \cite{hundt2019sharpdarts}, with amplitude decay factor of $0.5$ per cycle and initial value of $0.025$.
For regularization we used cutout  \cite{devries2017improved} with a length of $16$, scheduled drop-path \cite{larsson2016fractalnet} of $0.2$, auxiliary towers \cite{szegedy2015going} after the last reduction cell with a weight of $0.4$, label smoothing \cite{szegedy2016rethinking_label_smooth} of $0.1$, AutoAugment~\cite{cubuk2018autoaugment} and weight decay of $3\cdot10^{-4}$. 

{\bf ImageNet.}
Our training architecture starts with stem cells that reduce the input image resolution from $224$ to $56$ ($3$ reductions), similar to \cite{liu2018darts}. We then stack $14$ cells: $12$ normal cells and $2$ reduction cells. The reduction cells are placed after the fourth and eighth normal cells. The normal cells start with $46$ channels, as the number of channels is doubled after each reduction cell. We trained the network, with a batch size of $1280$ for $250$ epochs, with one cycle of power cosine learning rate, weight decay of $10^{-4}$ and nesterov-momentum of $0.9$.
We add an auxiliary loss after the last reduction cell with a weight of $0.4$.
During training, we normalize the input image and crop it with a random cropping factor in the range of $0.08$ to $1$. In addition we use auto-augment and horizontal flipping. 
During testing, we resize the input image to the size of $256$x$256$ and applying a fixed central crop to the size of $224$x$224$.

{\bf Additional datasets.}
Our additional classification datasets consist of CINIC-10 \cite{darlow2018cinic}, CIFAR-100 \cite{cifar100}, FREIBURG \cite{Freiburg}, SVHN \cite{SVHN} and FashionMNIST \cite{fashionMnist}. Their training scheme was similar to the one used for CIFAR-10, described at \ref{cifar10_training}, with some minor adjustments and modifications.
For the FREIBURG dataset, we resized the original images from $256$x$256$ to $96$x$96$ and used a batch size of $16$. For CINIC-10, we trained the network for $800$ epochs instead of $1500$, since this dataset is much larger then CIFAR-10. For Fashion-MNIST we edited the learned augmentations regime to fit a dataset of grayscale images.

\section{Cells Learned by XNAS}
\begin{figure}[H]
  \begin{subfigure}{}
    \centering
    \includegraphics[width=0.46\linewidth,height=7cm]{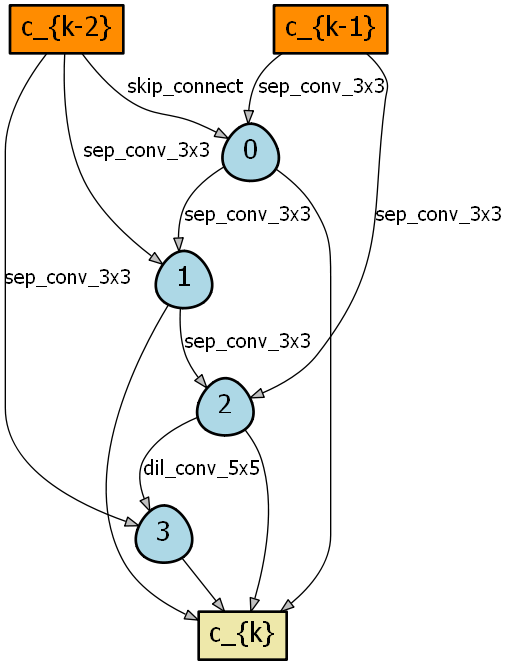}
  \end{subfigure}\qquad
  \begin{subfigure}{}
    \centering 
    \includegraphics[width=0.54\linewidth,height=7cm]{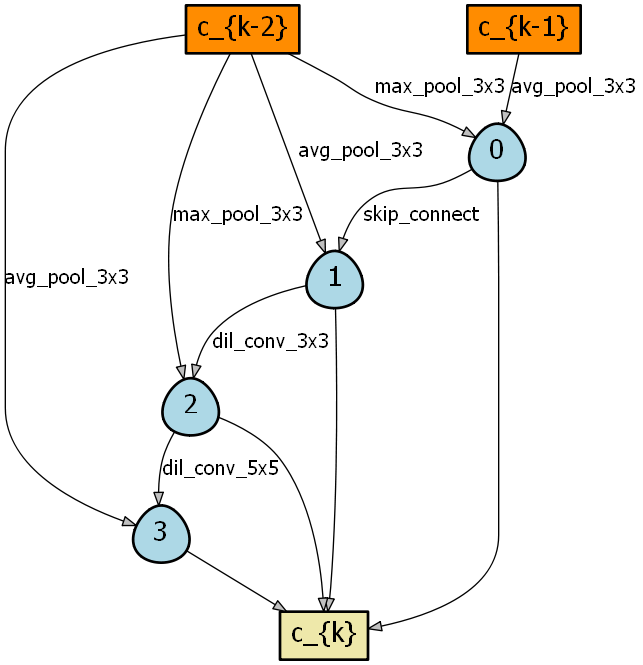}
  \end{subfigure}
\caption{XNAS learned normal and reduction cells on CIFAR-10.}  
 \label{fig:cells}
\end{figure}

Figure \ref{fig:cells} presents the cells learned by XNAS on CIFAR-10. 

{\bf{Cell depth:}}
When comparing the cell in Figure \ref{fig:cells} to other reported NAS cells \cite{noy2019asap,liu2018darts,Real18Regularized,NASNET,snas,PNAS}, it is clear visually that XNAS cell is "deeper" in some sense.

We wish to define a metrics for a cell "depth". A cell $C_{k}$ contains four nodes. Each node can be connected to the previous cell's nodes or to the two previous cells' outputs $C_{k-2}$, $C_{k-1}$. Let us index the previous cells $C_{k-2}$, $C_{k-1}$ and the cell's four nodes as $0,\dots,5$ respectively.

Define the depth of each connection in a cell as the index of the node (or previous cell) it came from. A simple metric for a cell depth can be the average depth of its inner connections.
Table \ref{cells_depth} presents the depth of XNAS and other NAS methods normal cells.

\begin{table}[H]
    \begin{center}
        \begin{tabular}{|l|c|}
        \hline
            Cell
            & \multicolumn{1}{|p{1.0cm}|}{\centering Depth } \\
            \hline
            SNAS \cite{snas} & $0.625$  \\ 
            PNAS \cite{PNAS} & $0.5$ \\
            Amoeba-A \cite{Real18Regularized} & $0.9$ \\ 
            NASNet \cite{NASNET} & $0.4$ \\   
            DARTS \cite{liu2018darts} & $0.625$ \\
            ASAP \cite{noy2019asap} & $0.875$ \\ 
            \hline
            XNAS & $1.375$ \\
           \hline
        \end{tabular}
    \end{center}
    \caption{XNAS and other NAS methods normal cell Depth.}  
    \label{cells_depth}
\end{table}
We can see from Table \ref{cells_depth} that the XNAS cell is much deeper then a typical NAS cell. This observation could provide a hint about the superior performance of the XNAS cell. Unlike most NAS methods, that usually produce shallow cells, XNAS cell utilizes better the possible degree of freedom of a NAS cell design, yielding a deeper and more complex architecture.

\medskip

\small
\end{document}